%% file: main.tex
\documentclass[acmtog, nonacm]{acmart}
\setcopyright{none}
\makeatletter
\let\@authorsaddresses\@empty
\def\@typeset@author@line{%
  \andify\@currentauthors\par\noindent
  \@currentauthors\def\@currentauthors{}%
  \def\@currentaffiliations{}}
\makeatother
\renewcommand\footnotetextcopyrightpermission[1]{}

\usepackage{cleveref}
\usepackage{multirow}
\usepackage{listings}

\input{preamble}

\begin{document}

%% Title
\title{RealMaster: Lifting Rendered Scenes into Photorealistic Video}

%% Authors
\author{Dana Cohen-Bar\textsuperscript{1,2}
\qquad Ido Sobol\textsuperscript{2,3}
\qquad Raphael Bensadoun\textsuperscript{2}
\qquad Shelly Sheynin\textsuperscript{2}
\\
Oran Gafni\textsuperscript{2}
\qquad Or Patashnik\textsuperscript{1}
\qquad Daniel Cohen-Or\textsuperscript{1}
\qquad Amit Zohar\textsuperscript{2}
\\ \\
\textsuperscript{1}Tel Aviv University
\qquad \textsuperscript{2}Reality Labs, Meta
\qquad \textsuperscript{3}Technion}
\affiliation{%
  \country{Israel}
}

%% Short author list for headers
\renewcommand{\shortauthors}{Cohen-Bar et al.}

%% CCS Concepts - generate at https://dl.acm.org/ccs/ccs.cfm
% <concept_id>10010147.10010178.10010224</concept_id>
% <concept_desc>Computing methodologies~Computer vision</concept_desc>

% \ccsdesc[500]{Computing methodologies~Computer vision}

%% Keywords
% \keywords{keyword1, keyword2, keyword3}

%% Teaser figure (optional - uncomment and add your image)
\input{figures/teaser}

%% Abstract
\input{0_abstract}

\maketitle

%% Paper sections
\input{1_intro}
\input{2_related}
\input{3_method}
\input{4_experiments}

\input{5_applications}

\input{6_conclusions}

%% Acknowledgments
\begin{acks}
We thank Ita Lifshitz and Daniel Garibi for their valuable contributions and support.
\end{acks}

%% Bibliography
\bibliographystyle{ACM-Reference-Format}
\bibliography{main}

\clearpage
\input{figures/figure_page_1}

\input{figures/figure_page_2}

%% Supplementary Material
\clearpage
\appendix
\begin{center}
  {\Huge\sffamily Supplementary Material}
\end{center}
\vspace{1em}
\input{supp}

\end{document}

%% file: preamble.tex
\usepackage{xcolor}
\definecolor{darkgreen}{RGB}{0,100,0} 

\usepackage{booktabs}       % Professional table formatting
\usepackage{adjustbox}      % For adjusting figure sizes
\usepackage{graphicx}       % For including graphics
\usepackage{subcaption}     % For subfigures

\newif\ifshowmarkup
\showmarkupfalse   % Change to \showmarkuptrue for markup

\crefname{figure}{Fig.}{Figs.}

\ifshowmarkup

  \newcommand{\danac}[1]{{\color{blue}[\textbf{Dana:} #1]}}
  \newcommand{\opc}[1]{{\color{darkgreen}[\textbf{OP:} #1]}}
  \newcommand{\dcc}[1]{{\color{red}[\textbf{DC:} #1]}}
  \newcommand{\amitc}[1]{{\color{teal}[\textbf{Amit:} #1]}}
  \newcommand{\ido}[1]{{\color{brown} #1}}
\else

  \newcommand{\danac}[1]{}
  \newcommand{\opc}[1]{}
  \newcommand{\dcc}[1]{}
  \newcommand{\amitc}[1]{}
  \newcommand{\ido}[1]{}
\fi

%% file: figures/teaser.tex
\begin{teaserfigure}
  \includegraphics[width=\textwidth,height=5cm,trim=3pt 3pt 3pt 3pt,clip]{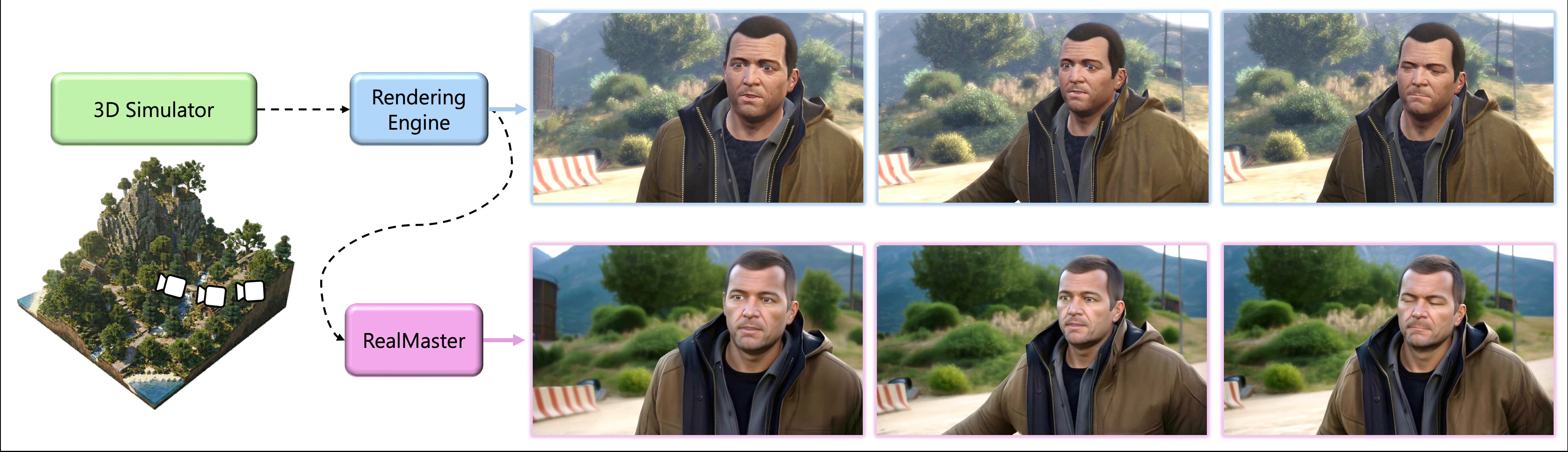}
  \caption{RealMaster lifts synthetic-looking rendered video into photorealistic video, faithfully re-realizing the original scene.}
  \Description{Teaser figure showing sim-to-real transformation.}
  \label{fig:teaser}
\end{teaserfigure}

%% file: 0_abstract.tex
\begin{abstract}
    State-of-the-art video generation models produce remarkable photorealism, 
    but they lack the precise control required to align generated content with 
    specific scene requirements. Furthermore, without an underlying explicit 
    geometry, these models cannot guarantee 3D consistency. Conversely, 3D 
    engines offer granular control over every scene element and provide native 
    3D consistency by design, yet their output often remains trapped in the 
    ``uncanny valley''. 
    Bridging this sim-to-real gap requires both \emph{structural precision}, 
    where the output must exactly preserve the geometry and dynamics of the input, 
    and \emph{global semantic transformation}, where materials, lighting, and 
    textures must be holistically transformed to achieve photorealism.
    We present RealMaster, a method that leverages video diffusion models 
    to lift rendered video into photorealistic video while maintaining 
    full alignment with the output of the 3D engine. To train this model, we 
    generate a paired dataset via an anchor-based propagation strategy, 
    where the first and last frames are enhanced for realism and propagated 
    across the intermediate frames using geometric conditioning cues. 
    We then train an IC-LoRA on these paired videos to distill the 
    high-quality outputs of the pipeline into a model that generalizes beyond 
    the pipeline's constraints, handling objects and characters that appear 
    mid-sequence and enabling inference without requiring anchor frames.
    Evaluated on complex GTA-V sequences, RealMaster significantly outperforms 
    existing video editing baselines, improving photorealism while preserving 
    the geometry, dynamics, and identity specified by the original 3D control.
\end{abstract}

%% file: 1_intro.tex
\section{Introduction}
\label{sec:intro}

Recent advancements in large-scale generative models have enabled the 
synthesis of video with extraordinary photorealism. However, these models remain 
difficult to steer with precision: they rely on text prompts or reference images 
rather than explicit 3D representations, limiting their capacity to control 
individual scene elements or guarantee geometric consistency across frames.

In contrast, traditional 3D engines offer precise user control and enforce 
geometric consistency by design. Yet, despite decades of progress in rendering, 
the \textit{sim-to-real} gap persists: synthetic outputs often retain a sterile 
appearance that lacks the high-frequency detail of real-world footage, 
often falling into the uncanny valley (see \cref{fig:teaser}, top).
Bridging this gap would enable a compelling new paradigm: using video 
diffusion models as a learned second-stage renderer atop fast 3D engines, 
combining the control of traditional graphics with the photorealism of 
generative models.

To bridge this gap, the task of sim-to-real translation aims to transform 
rendered video into photorealistic sequences. 
A natural approach is to leverage recent advances in video editing, where 
large-scale generative models have demonstrated impressive capabilities in 
modifying video content while preserving temporal coherence. However, sim-to-real 
translation poses a fundamentally different challenge than standard video editing. 
Unlike typical editing tasks, which involve local modifications or global 
stylization, sim-to-real requires simultaneously satisfying two seemingly 
conflicting objectives: \emph{structural precision}, where the output must exactly 
preserve the input's geometry, motion, and dynamics down to fine details; and 
\emph{global semantic transformation}, where materials, lighting, and textures 
must be holistically transformed to achieve true photorealism. Because the input 
is already near-photorealistic, details cannot be abstracted away as in 
conventional style transfer; the model must preserve fine details while adding 
the high-frequency nuances that characterize real-world footage. In practice, 
we find that existing video editing methods struggle with this tension. When 
applied to sim-to-real, they either fail to recognize the synthetic nature of 
the input and leave it largely unchanged, or they change too much and fail to 
preserve important details from the original.

In this work, we present RealMaster, a method for sim-to-real video translation.
Specifically, we train a model that lifts rendered video into photorealistic video while preserving the underlying scene structure and dynamics. A central component of our approach is a sparse-to-dense propagation strategy that constructs high-quality training supervision directly from rendered sequences. Given a rendered video, we first edit the first and last frames to serve as photorealistic visual anchors. We then propagate their appearance across the sequence using a conditional video model guided by edge cues, producing a photorealistic video that remains aligned with the original rendered input. This process yields paired rendered–photorealistic video data. We then train an IC-LoRA on these video pairs, distilling the behavior of the propagation pipeline into a model that generalizes beyond its limitations and can directly perform the sim-to-real task at inference time. By leveraging the foundation model as a strong prior, the network learns to discount imperfections in the synthetic data and produce high-quality outputs that remain faithful to the input rendered video.

% In this work, we present RealMaster, a method for sim-to-real video translation. 
% Specifically, we train a model that lifts rendered video 
% into photorealistic video while preserving the underlying scene structure and 
% dynamics. A central component of our approach is a sparse-to-dense propagation 
% strategy that constructs high-quality training supervision directly from 
% rendered sequences. Given a rendered video, we first edit the first and last frames to serve as photorealistic visual anchors. We then propagate their appearance across the sequence using a conditional video model guided by edge cues, producing a photorealistic video that remains aligned with the original rendered input. We then train 
% an IC-LoRA on the rendered and photorealistic video pairs, allowing the model 
% to learn the sim-to-real mapping while preserving the base model's prior. By 
% leveraging the foundation model as a strong prior, the network learns to discount 
% imperfections in the synthetic data and produce high-quality outputs that remain 
% faithful to the input rendered video.

\input{figures/method}

We evaluate the effectiveness of RealMaster through extensive experiments on diverse sequences from the GTA-V virtual environment. This setting provides a challenging testbed due to its complex lighting transitions, high-speed motion, intricate geometric details, and the presence of multiple interacting characters. As shown in \cref{fig:teaser}, RealMaster produces photorealistic videos that preserve the structure and dynamics of the source scenes under these challenging conditions. Our quantitative and qualitative results further demonstrate that RealMaster significantly outperforms state-of-the-art video editing baselines in both preservation of the input and photorealism, successfully resolving the trade-off between structural precision and global transformation that limits existing methods.

%% file: figures/method.tex
\begin{figure*}[!t]
    \centering
    \includegraphics[width=0.9\linewidth,trim=1pt 1pt 1pt 4pt,clip]{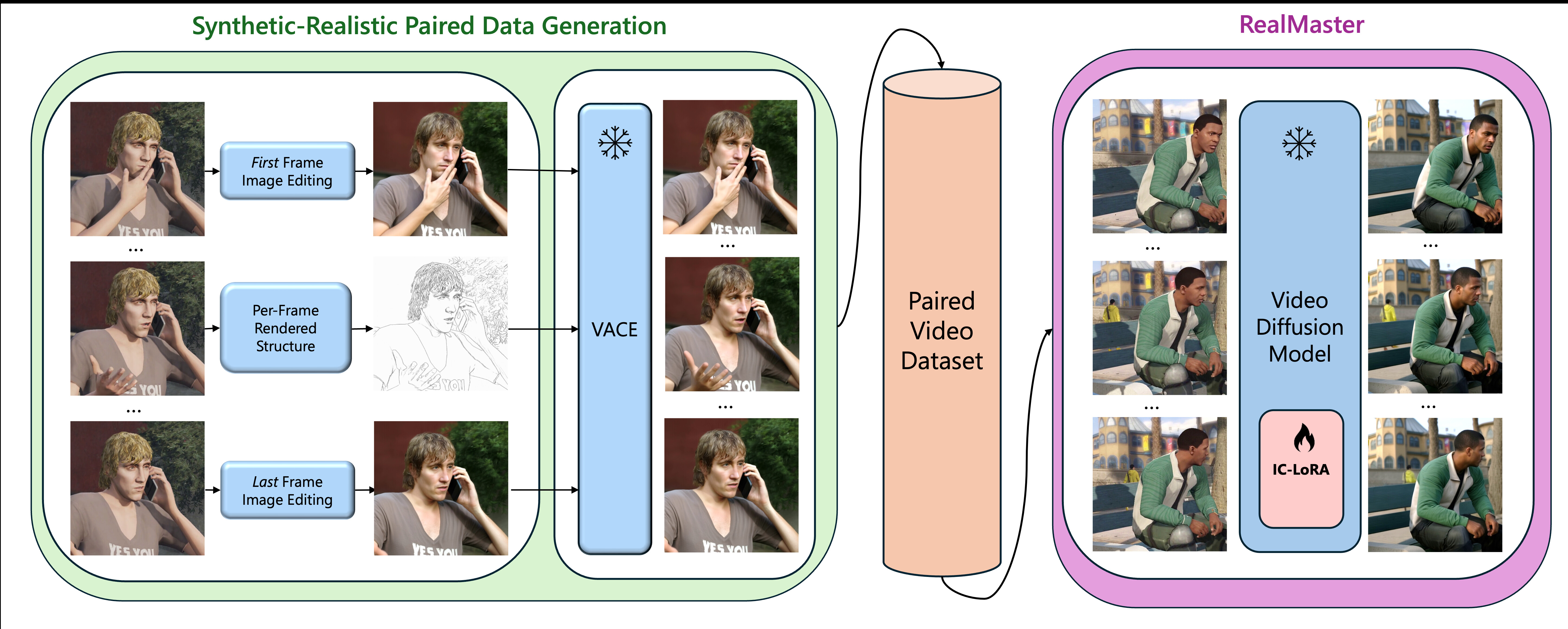}
    \caption{\textbf{Overview of RealMaster.}
    Our method consists of two stages: (1) \textbf{Synthetic-to-Realistic Data Generation}: Given a synthetic video, we edit sparse keyframes and propagate their appearance across the sequence using VACE, conditioned on edge maps from the input video, to create paired synthetic–realistic training data. (2) \textbf{Model Training}: We fine-tune an IC-LoRA over a text-to-video diffusion model on the paired data, enabling direct sim-to-real video translation at inference time.}
    \label{fig:method}
\end{figure*}

%% file: 2_related.tex
\section{Related Work}
\label{sec:related}

\subsection{Sim-to-Real Translation}
The mapping of rendered content to photorealistic domains is fundamentally distinct 
from artistic style transfer. This problem was first explored in classical example-based 
synthesis, most notably the Image Analogies framework~\citep{hertzmann2001image}, 
which introduced non-parametric mappings between paired images to transfer complex 
textures. Building on this logic, \citet{johnson2011cg2real} developed CG2Real, 
leveraging large-scale image retrieval to inject real-world statistics into 
computer-generated images. While these early methods established the importance of 
data-driven anchors, they relied on manual feature matching and lacked the robust 
generative priors inherent in modern foundation models.

Subsequent efforts shifted toward deep generative architectures that replace manual 
matching with learned representations. Early image-to-image translation via 
conditional GANs~\citep{isola2017image, zhu2017unpaired, yi2017dualgan, 
liu2017unsupervised} refined these analogies into global mappings but often 
struggled with the photometric precision required for sim-to-real tasks. 
To bridge this gap, \citet{chen2017learning} and \citet{richter2021enhancing}
demonstrated that incorporating engine-specific G-buffers, including depth and 
surface normals, significantly improves geometric grounding in complex sequences.
Recent work~\citep{wang2025zero} explores zero-shot diffusion-based realism enhancement for synthetic videos, demonstrating promising results on egocentric driving data. In this work, we study sim-to-real translation for videos containing rendered humans, where preserving character identity and articulated motion introduces additional challenges compared to primarily rigid-object scenes.

\subsection{Video Generation and Controllability}

Recent breakthroughs in diffusion-based generative models~\citep{ho2020denoising, song2021score} have redefined video synthesis. Foundation models such as Stable Video Diffusion~\citep{blattmann2023stable}, Gen-2~\citep{esser2023structure}, Lumiere~\citep{bar-tal2024lumiere}, CogVideoX~\citep{yang2024cogvideox}, MovieGen~\citep{polyak2024movie}, Wan~\citep{wan2025wan} and LTX-2~\citep{hacohen2026ltx2} produce high-resolution, cinematic sequences. 

In parallel to these advances in video generation, a growing body of work studies controllability through explicit conditioning. 
ControlNet~\citep{zhang2023adding} introduced a paradigm for conditioning image diffusion models on spatial control signals such as depth, edges, and human pose. 
Subsequent work extends structural conditioning to video diffusion by providing these signals across time, including depth-conditioned generation~\citep{luo2023videofusion}, temporally sparse constraints~\citep{guo2024sparsectrl}, and training-free ControlNet-style control for text-to-video~\citep{zhang2024controlvideo}.

Complementary to structural conditioning, exemplar-based approaches use in-context visual examples to guide generation. 
In-Context LoRA~\citep{huang2024context} demonstrates this for text-to-image diffusion transformers, showing that the model can learn to leverage structured exemplars provided in the context during generation, and that this capability can be further strengthened through lightweight fine-tuning.

\subsection{Video Editing}

Diffusion-based video generation models have been extended to video editing through two main paradigms. 
Early work largely operates in a zero-shot manner, enabling text-guided manipulation without requiring task-specific paired training data~\citep{wu2023tune, qi2023fatezero, geyer2023tokenflow, chen2023vid2vidzero, liu2024videop2p, singer2024video, yang2023rerender, cong2023flatten}. 
In contrast, more recent approaches leverage large-scale training to support general-purpose video editing capabilities across a wide range of edits~\citep{molad2023dreamix, qin2023instructvid2vid, polyak2024movie, jiang2025vace, lucyedit, bai2025ditto}.

A complementary line of work focuses on first-frame editing followed by propagation~\citep{ku2024anyv2v, ceylan2023pix2video, ouyang2024codef, ouyang2024i2vedit}, where sparse edits are transferred across time using a conditional video model. This paradigm is most closely related to our approach, as it similarly aims to maintain temporal coherence while applying targeted appearance changes.

However, despite strong performance on creative edits, existing video editing methods struggle on sim-to-real translation. When applied to rendered videos, they either fail to recognize the synthetic appearance and thus produce minimal changes, or they introduce large visual edits that fail to preserve the underlying scene structure and character identity. This limitation highlights a fundamental tension in sim-to-real translation: the task requires both global appearance transformation and strict input preservation—objectives that current video editing methods struggle to optimize jointly.

% \danac{need to add: wan, vace, ltx-video, controlvideo, I2VEdit, Pix2Video, dynVFX}

%% file: 3_method.tex
\section{Method}

\label{sec:method}

Our goal is to transform rendered 3D engine outputs into photorealistic 
video while preserving the underlying scene structure and dynamics.
We achieve this through a two-stage approach: first, we construct 
high-quality paired training data via a data generation pipeline. Then, we 
train an IC-LoRA adapter that distills the data generation pipeline behavior into a model with
improved generalization beyond the pipeline's inherent constraints.
An overview of our method is shown in \cref{fig:method}.

\subsection{Data Generation Pipeline}
\label{subsec:data_generation}

A central challenge in sim-to-real video translation is the lack of paired data aligning rendered engine outputs with corresponding photorealistic videos. To address this, we develop a pipeline that directly constructs photorealistic counterparts from rendered videos.

Image-based sim-to-real translation is more mature and reliable than its video equivalent. We therefore adopt a sparse-to-dense strategy: we edit a small set of keyframes using an image editing model to establish the target photorealistic appearance, and then propagate this appearance to intermediate frames using a video model with structural conditioning.

\paragraph{\textbf{Keyframe Enhancement.}}
Given a rendered video sequence, we first translate the first and last 
frames into the photorealistic domain using an off-the-shelf image editing 
model \citep{wu2025qwenimagetechnicalreport}. These enhanced keyframes serve as appearance anchors that define the 
target photorealistic look for the full sequence.

\paragraph{\textbf{Edge-Based Keyframe Propagation.}}
To propagate keyframe appearance to intermediate frames, we utilize VACE~\citep{jiang2025vace}, a video generative model that conditions generation on reference frames and structural signals.

Specifically, we extract edge maps from the input video and use VACE to generate the full video conditioned on the photorealistically edited keyframes and the corresponding edge maps. Edge conditioning anchors generation to the input’s structure and motion, allowing VACE to propagate the keyframe appearance while preserving scene layout and dynamics across intermediate frames.

% \paragraph{\textbf{Edge-Based Conditioning Schedule.}}
% To propagate the keyframe appearance to intermediate frames, we build upon VACE~\danac{add citation}, a video generative model that supports conditioning on reference frames together with structural signals such as edge maps. We leverage this capability by injecting edge-based structural guidance during the diffusion process, motivated by the observation that different denoising stages benefit from different types of conditioning.

% In the early diffusion timesteps, where the model establishes the global layout, edge conditioning provides coarse structural guidance that anchors the generated content to the underlying scene geometry. In the later timesteps, where the model refines fine details, edge conditioning preserves high-frequency information such as object boundaries and texture patterns, which is crucial for maintaining identity. This coarse-to-fine use of edge guidance aligns with the progressive nature of diffusion and helps address the core challenge of sim-to-real translation: preserving scene structure and identity while enabling realistic appearance transformation.

% \paragraph{\textbf{Quality Filtering.}}
% Although the image editing model yields high-quality photorealistic keyframes, it sometimes modifies character identity. We therefore filter generated pairs using ArcFace~\citep{deng2019arcface} identity similarity scores, retaining only samples that preserve identity to provide a cleaner training signal.
\input{figures/qualitative_results}

\subsection{Model Training}
\label{subsec:model_training}

We train a lightweight LoRA adapter that distills our data generation pipeline into a single model for sim-to-real video translation. Specifically, we adopt an IC-LoRA architecture on top of a pre-trained text-to-video diffusion backbone. During training, we concatenate clean reference tokens from the rendered input video with noisy  tokens and optimize the model to denoise toward the corresponding photorealistic target. Training is lightweight, requiring only a small paired dataset and a few hours of fine-tuning on a single GPU.

At inference time, the resulting model avoids several constraints imposed by the pipeline. First, the pipeline requires access to both the first and last frames of a sequence, which makes streaming or autoregressive generation impractical. Second, because edits are anchored to sparse keyframes, the pipeline struggles to preserve the appearance and identity of objects and characters that emerge mid-sequence. Third, the image editing model can over-edit anchor frames, causing deviations from the input scene.

Overall, the trained model removes these inference-time constraints, enabling temporally coherent sim-to-real translation while preserving scene structure and character identity.

\vspace{-10pt}
\subsection{Implementation Details}
\label{subsec:impl_details}
For data generation, we sample clips from the SAIL-VOS~\citep{hu2019sailvos}
training set, upsampling them from 8\,fps to 16\,fps by repeating each
frame to obtain 81-frame sequences at $800\times1200$ resolution. We edit the keyframes using Qwen-Image-Edit~\citep{wu2025qwenimagetechnicalreport} and propagate their appearance to intermediate frames using VACE~\citep{jiang2025vace} conditioned on edge maps. To improve identity consistency in the generated pairs, we filter out clips whose minimum ArcFace~\citep{deng2019arcface} cosine similarity between faces detected in the source and edited videos falls below 0.4. This process yields a training set of 1,216 clips.

For model training, we fine-tune Wan2.2 T2V-A14B~\citep{wan2025wan} using a LoRA adapter with a rank of 32. Following IC-LoRA~\citep{huang2024context}, we encode the rendered input as clean reference tokens with their timestep fixed to $t{=}0$, sharing positional encoding with the noisy tokens being denoised.
%Full hyperparameters are provided in the supplementary material.

%% file: figures/qualitative_results.tex
\begin{figure*}[t]
    \centering
    \includegraphics[width=\linewidth]{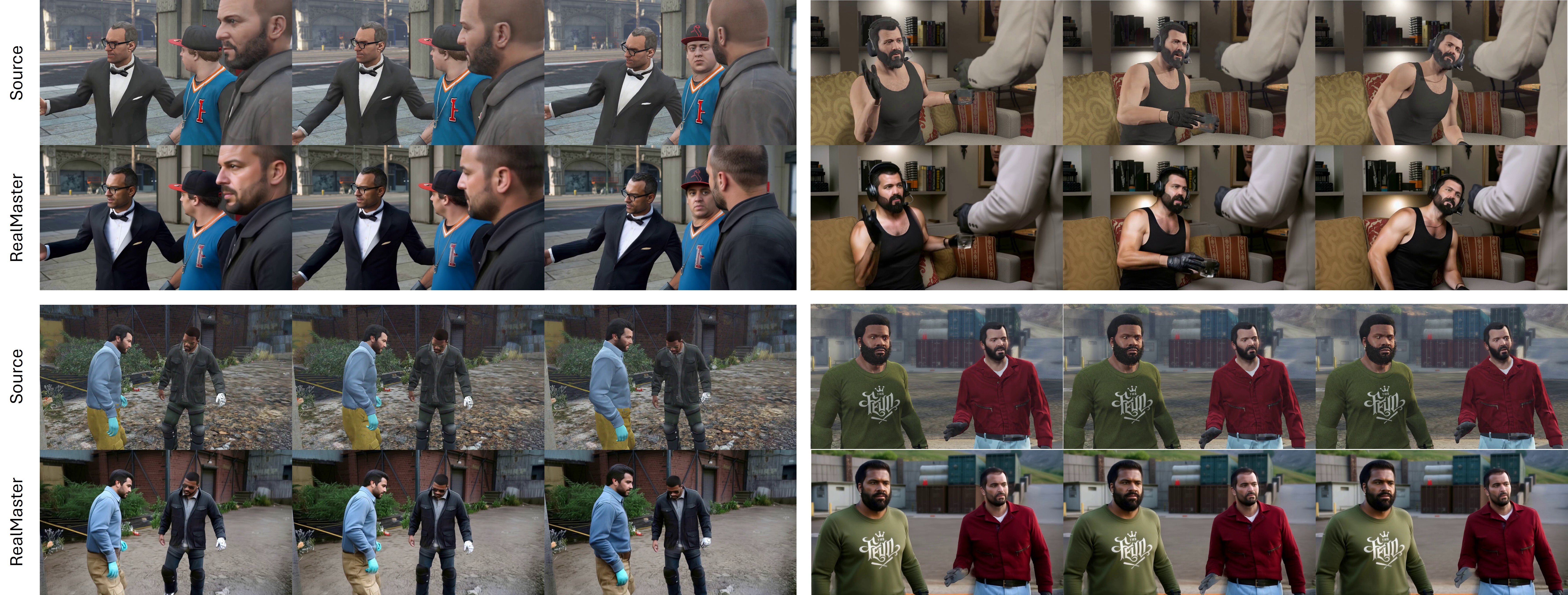}
    \caption{\textbf{Qualitative Results.}
    We show representative GTA-V video sequences together with their edits produced by our method. These translated sequences demonstrate our method’s ability to produce photorealistic video while maintaining strict temporal coherence. Note the consistent appearance of materials, lighting, and fine details across frames. Best viewed zoomed in.}
    \label{fig:qualitative_results}
\end{figure*}

%% file: 4_experiments.tex
\input{figures/comparison}
\section{Experiments}
\label{sec:experiments}

We perform a series of experiments to evaluate RealMaster. First, we compare our approach against strong baselines for video editing and sim-to-real translation. Second, we conduct ablation studies to assess the impact of key design choices in our approach.

\subsection{Experimental Setup}
\label{subsec:experimental_setup}
We use a subset of 100 clips sampled uniformly from the SAIL-VOS validation set for our experiments. SAIL-VOS is recorded at 8\,fps, and we upsample it to 16\,fps by repeating each frame. The validation set contains diverse GTA-V scenarios featuring multiple interacting characters and visually complex scenes with many objects.
We evaluate all methods using both automatic metrics and human evaluation. Both assess key aspects such as photorealism, input preservation, and temporal consistency.

\paragraph{\textbf{Automatic Metrics.}}
We evaluate identity consistency, structure preservation, realism, and temporal consistency using complementary automatic metrics.
To measure identity consistency, we compute the mean ArcFace similarity between faces detected in the input and edited videos. Specifically, we uniformly sample five frames per video, match the detected faces between the input and edited frames, and report the average cosine similarity of their ArcFace embeddings.
We assess structure preservation by measuring the $\ell_2$ distance between DINO features extracted over all frames of the input and edited videos. This metric captures high-level semantic and structural consistency between the rendered input and the photorealistic output.

For realism assessment, we use GPT-4o to rate the photorealism of edited frames on a scale from 1 to 10. For each video, we uniformly sample five frames and report the average score. We conduct this evaluation under two settings: (i) GPT-RS$_{\text{no-ref}}$, where only the edited frame is provided to GPT-4o, and (ii) GPT-RS$_{\text{with-ref}}$, where the corresponding input frame is provided alongside the edited frame. This allows us to assess realism both in isolation and relative to the rendered input. 

To evaluate temporal consistency, we adopt the Temporal Flickering and Motion Smoothness metrics from VBench~\citep{vbench}. Temporal Flickering measures frame-to-frame visual instability, capturing abrupt appearance changes across consecutive frames, while Motion Smoothness assesses the coherence of motion over time. 

% \paragraph{\textbf{User Study.}}
% For human evaluation, we conduct a pairwise preference study comparing RealMaster with each baseline. In each trial, participants are shown a reference video (the original rendered input) and two corresponding AI-converted videos (Video A and Video B), all playing automatically and in sync. For each comparison, participants answer three questions: (i) Realism — which video makes the game footage look more real; (ii) Faithfulness — which video better preserves the structure, motion, and identity of the original video; and (iii) Visual Quality — which video is more aesthetically pleasing. 

\paragraph{\textbf{Baselines.}}
We compare our method against three strong video editing methods: Runway-Aleph~\cite{runway2025aleph}, LucyEdit~\citep{lucyedit} and Editto~\citep{bai2025ditto}. Among these, Editto is explicitly trained for sim-to-real translation using synthetic-real pairs.

% \paragraph{\textbf{Dataset.}}
% We evaluate on a curated benchmark of GTA-V sequences spanning diverse scenarios including urban driving, pedestrian scenes, and complex lighting transitions.
% TODO: Add specific details about the benchmark: number of sequences, duration, resolution, etc.

\subsection{Qualitative Results}

\label{subsec:qualitative_results}

% As shown in \autoref{fig:qualitative_results}, our method successfully transforms synthetic renderings into photorealistic video. The output preserves fine-grained details such as lighting gradients, material properties, and high-frequency textures while maintaining strict alignment with the underlying 3D geometry.
As shown in \cref{fig:qualitative_results}, our method transforms rendered videos toward the photorealistic domain. The results preserve scene structure and
motion, as well as character identity and appearance, while improving material
and lighting realism.

These improvements are demonstrated in dynamic, cluttered scenes with multiple interacting characters, camera motion, and frequent occlusions, showing that the method successfully enhances realism despite challenging conditions that stress both structural precision and global semantic transformation.

\input{figures/user_study_table}

\cref{fig:comparison} presents a qualitative comparison with the baselines.
Runway-Aleph can improve realism but shifts object colors and does not preserve
character identity. LucyEdit pushes the output toward a more game-like
appearance than the input and alters many details of the original scene.
Editto, despite training on paired synthetic--real data, deviates significantly
from the content of the original scene. In contrast, RealMaster preserves structure and identity while substantially improving visual realism.

\subsection{Quantitative Comparison}
\label{subsec:quantitative_comparison}

As shown in \autoref{tab:comparison_metrics}, our method outperforms all baselines on most evaluated metrics. It achieves the highest scores on both GPT-RS$_{\text{no-ref}}$ and GPT-RS$_{\text{ref}}$, indicating superior photorealism both in isolation and relative to the rendered input. It also obtains the best ArcFace score and the lowest DINO score, demonstrating improved preservation of character identity and structural fidelity.

For temporal consistency, our method is competitive with the strongest baselines. It matches the best Temporal Flickering score and achieves comparable Motion Smoothness. While LucyEdit attains a slightly higher
Motion Smoothness score, it does so by blurring the video, which
reduces high-frequency detail and can inflate smoothness metrics while
degrading structural precision.

Overall, these results indicate that our method provides a better balance between photorealism, identity and structure preservation, and temporal consistency for sim-to-real video translation.

\input{tables/comparison_metrics}

\subsection{User Study}
\label{subsec:user_study}
To further validate our results, we conduct a user preference study comparing our method against the three baselines. In each trial, participants view the original rendered input together with two enhanced outputs (RealMaster vs.\ one baseline) and answer three questions assessing realism, faithfulness to the original video, and overall visual quality. In total, we collect 675 pairwise comparisons from 45 participants across the benchmark. As shown in \cref{fig:user_study}, our method is preferred over all baselines across all three metrics.

\input{figures/ablation_datagen}
\input{figures/ablation_pipeline_vs_model}

% As shown in \autoref{fig:user_study}, our method is consistently preferred across all evaluation criteria, confirming that the improvements observed in quantitative metrics translate to perceptually meaningful gains. \amitc{repeats previous paragraph}
% TODO: Add specific preference percentages and statistical significance.

\subsection{Ablation Studies}
\label{subsec:ablations}

We conduct ablation studies to compare alternative design choices in our data generation pipeline and to quantify the additional gains from training a model on the generated data. For each sequence, we edit the first and last frames and explore different strategies for propagating their appearance to intermediate frames using VACE. Specifically, we compare: (i) editing additional anchor frames at regular intervals (one every 0.5 seconds) and conditioning VACE on these anchors, (ii) conditioning VACE on depth maps, and (iii) conditioning VACE on edge maps (our default pipeline). Finally, we compare these pipeline variants to our full method (RealMaster), which trains a LoRA-adapted model on data generated with edge-based propagation.

In \cref{fig:ablation_datagen}, we show qualitative comparisons of these propagation variants. \textit{Multiple anchors} often introduce flickering, as inconsistencies across independent image edits are amplified during interpolation. \textit{Depth} provides coarse geometric guidance but can miss high-frequency cues important for identity and facial expressions. \textit{Edges} more reliably preserve object boundaries and fine facial details, improving structural precision in the generated training pairs..

In \cref{fig:ablation_pipeline_vs_model}, we compare inference with our trained model to direct use of the data generation pipeline. The model generalizes to cases where the pipeline fails, such as when an object first appears between the two boundary anchors, for which the pipeline has no appearance supervision. It also better preserves the appearance of the rendered input and avoids changes that are sometimes overly aggressive from the image editing model.

\autoref{tab:ablation_metrics} quantifies these trends. \textit{Multiple Anchors}
and \textit{Depth} perform worse on ArcFace and DINO, which reflect preservation
of identity and scene structure. \textit{Edges}, which is the pipeline used to
generate training pairs, yields the strongest pipeline scores on these metrics
while maintaining stable Temporal Flickering and Motion Smoothness.
Training \textit{RealMaster} on the generated data pairs further improves all
metrics, with the largest gains in structure and temporal consistency.
% TODO: Add discussion of ablation metrics.

\input{tables/ablation_metrics}

% \subsection{Additional Results}
% \label{subsec:additional_results}
% 
% % TODO: Optionally add section for physics and effects using prompts (rain, snow, etc.)

%% file: figures/comparison.tex
\begin{figure*}[!t]
    \centering
    \includegraphics[width=\textwidth]{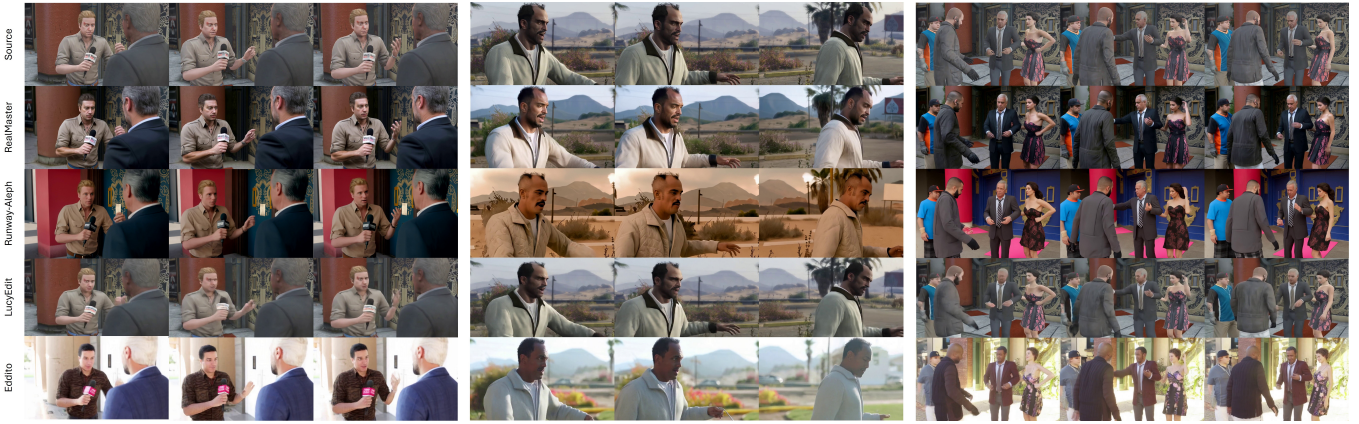}
    \caption{\textbf{Qualitative comparison with baseline methods.}
    We compare our method against Runway-Aleph, LucyEdit, and Editto on three videos from the benchmark. The baselines either alter the original scene content, leading to identity drift and color shifts, or fail to produce sufficiently photorealistic results. In contrast, our method preserves scene structure and identity while improving the photorealism.}
    \label{fig:comparison}
\end{figure*}

%% file: figures/user_study_table.tex
\begin{figure}[t]
  \centering
  \small
  \begin{adjustbox}{max width=\linewidth}
    \input{tables/user_study_tabular}

  \end{adjustbox}
  \Description{Table reporting the percentage of trials where participants
    preferred RealMaster over each baseline for realism, faithfulness, and
    visual quality.}
  \caption{\textbf{User study.} We report the percentage of trials where participants preferred RealMaster over each baseline for realism, faithfulness to the original video, and overall visual quality.}
  \label{fig:user_study}
\end{figure}

%% file: tables/user_study_tabular.tex
\begin{tabular}{lccc}
  \toprule
  Method & Realism & Faithfulness & Visual Quality \\
  \midrule
  vs.\ Editto & 63\% & 94\% & 78\% \\
  vs.\ LucyEdit & 93\% & 85\% & 93\% \\
  vs.\ Runway-Aleph & 64\% & 88\% & 70\% \\
  \midrule
  Overall & 73\% & 89\% & 80\% \\
  \bottomrule
\end{tabular}

%% file: tables/comparison_metrics.tex
\begin{table}[h]
    \centering
    \caption{\textbf{Quantitative comparison against baselines.} 
    We compare our method against baseline approaches using automatic metrics on our benchmark.}
    \label{tab:comparison_metrics}
    \footnotesize
    \setlength{\tabcolsep}{1pt}
    \renewcommand{\arraystretch}{0.9}
    \resizebox{\columnwidth}{!}{%
    \begin{tabular}{lcccccc}
        \toprule
        \textbf{Method} 
        & \scriptsize{\textbf{GPT-RS$_{\text{no-ref}}$}}$\uparrow$ 
        & \scriptsize{\textbf{GPT-RS$_{\text{ref}}$}}$\uparrow$ 
        & \scriptsize{\textbf{ArcFace}}$\uparrow$ 
        & \scriptsize{\textbf{DINO}}$\downarrow$ 
        & \scriptsize{\textbf{Temp. Flicker}}$\uparrow$ 
        & \scriptsize{\textbf{Mot. Smooth.}}$\uparrow$ \\
        \midrule
        Editto      
        & 5.104 & 3.838 
        & 0.204 
        & 41.79 
        & 0.972 & 0.972 \\
        Runway-Aleph 
        & 4.98 & 5.33 
        & 0.300 
        & 38.04 
        & \textbf{0.976} & 0.972 \\
        LucyEdit    
        & 3.48 & 4.20 
        & 0.375
        & 36.68
        & \textbf{0.976} & \textbf{0.986} \\
        \midrule
        \textbf{RealMaster} 
        & \textbf{5.296} & \textbf{7.33} 
        & \textbf{0.473} 
        & \textbf{30.28} 
        & \textbf{0.976} & 0.973 \\
        \bottomrule
    \end{tabular}%
    }
\end{table}

%% file: figures/ablation_datagen.tex
\begin{figure}[h]
    \centering
    \setlength{\tabcolsep}{0pt}
    \renewcommand{\arraystretch}{0}
    \begin{tabular}{@{}c@{\hspace{4pt}} c@{}c@{}c@{}}
        & \small Frame 1 & \small Frame 2 & \small Frame 3 \\[2pt]
        \smash{\rotatebox{90}{\small\hspace{8pt} Source}} &
        \includegraphics[width=0.31\linewidth]{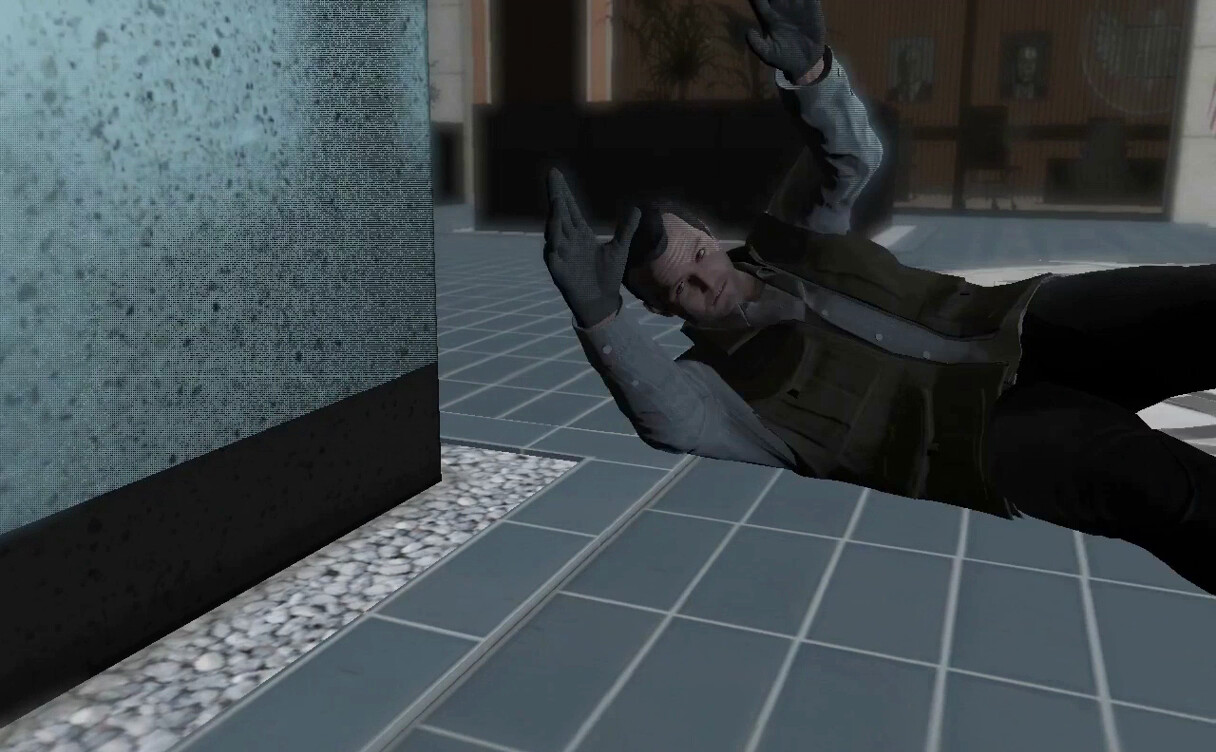} &
        \includegraphics[width=0.31\linewidth]{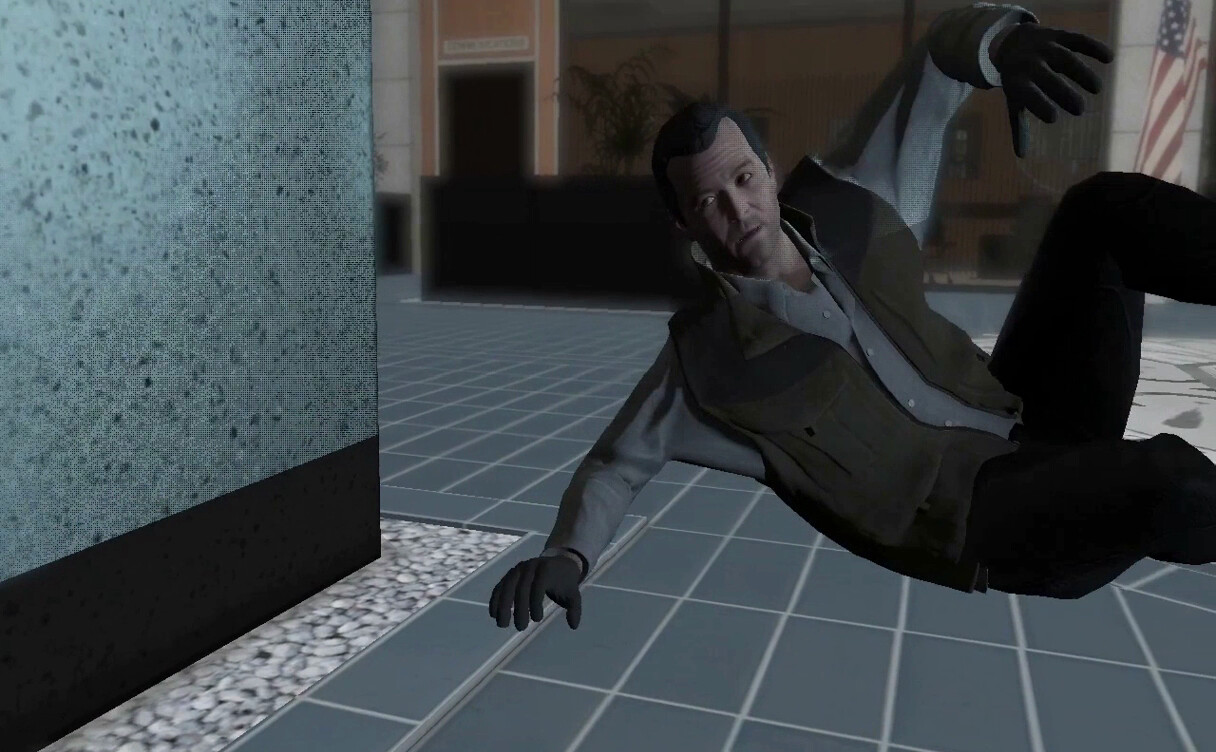} &
        \includegraphics[width=0.31\linewidth]{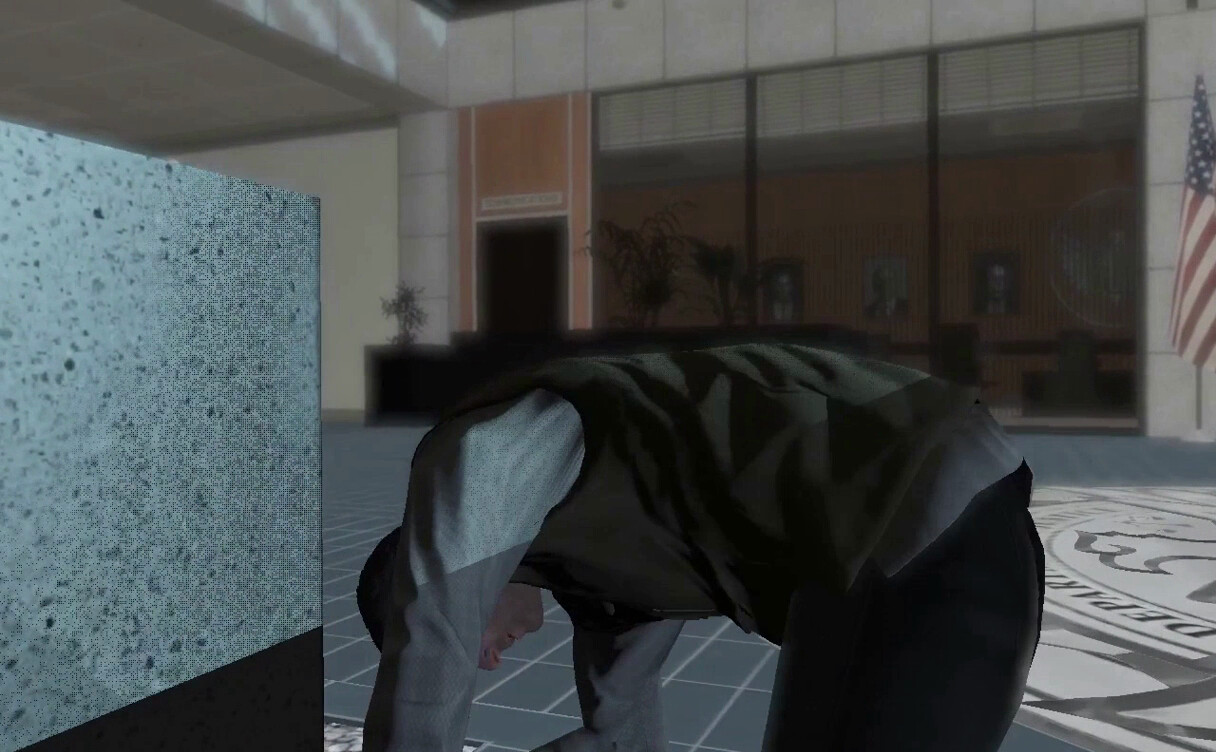} \\
        \smash{\rotatebox{90}{\small Edges (ours)}} &
        \includegraphics[width=0.31\linewidth]{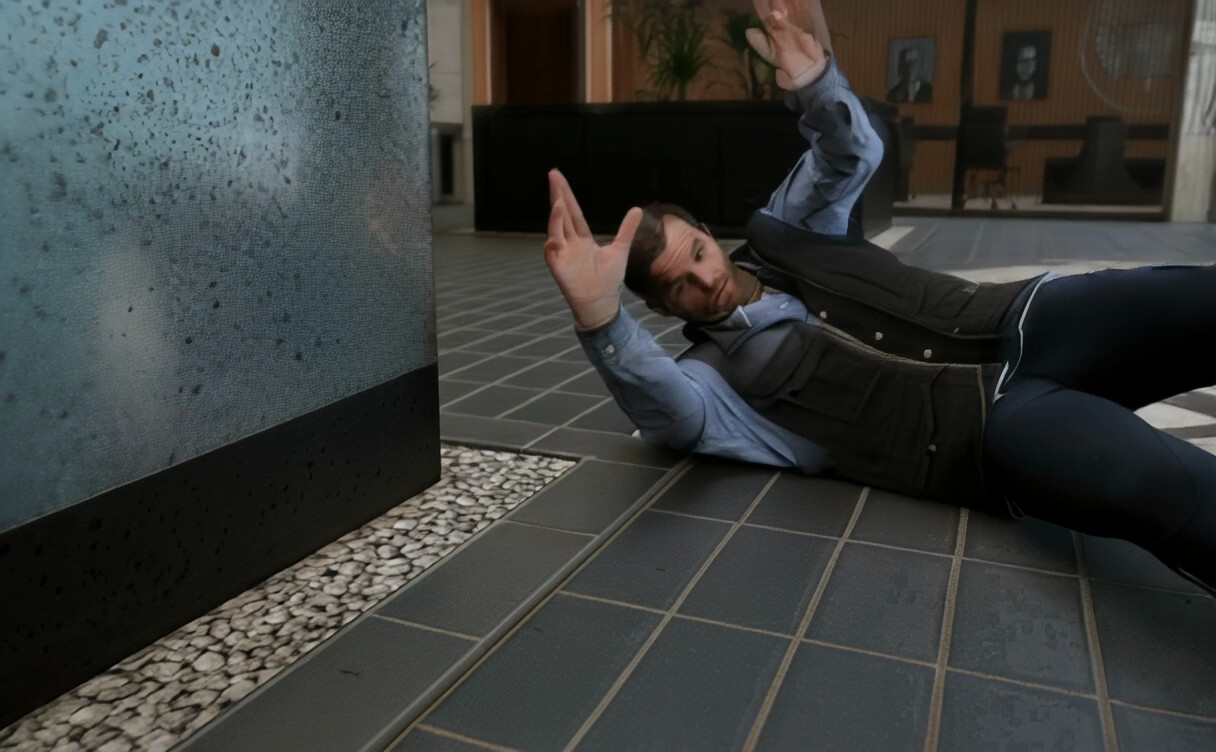} &
        \includegraphics[width=0.31\linewidth]{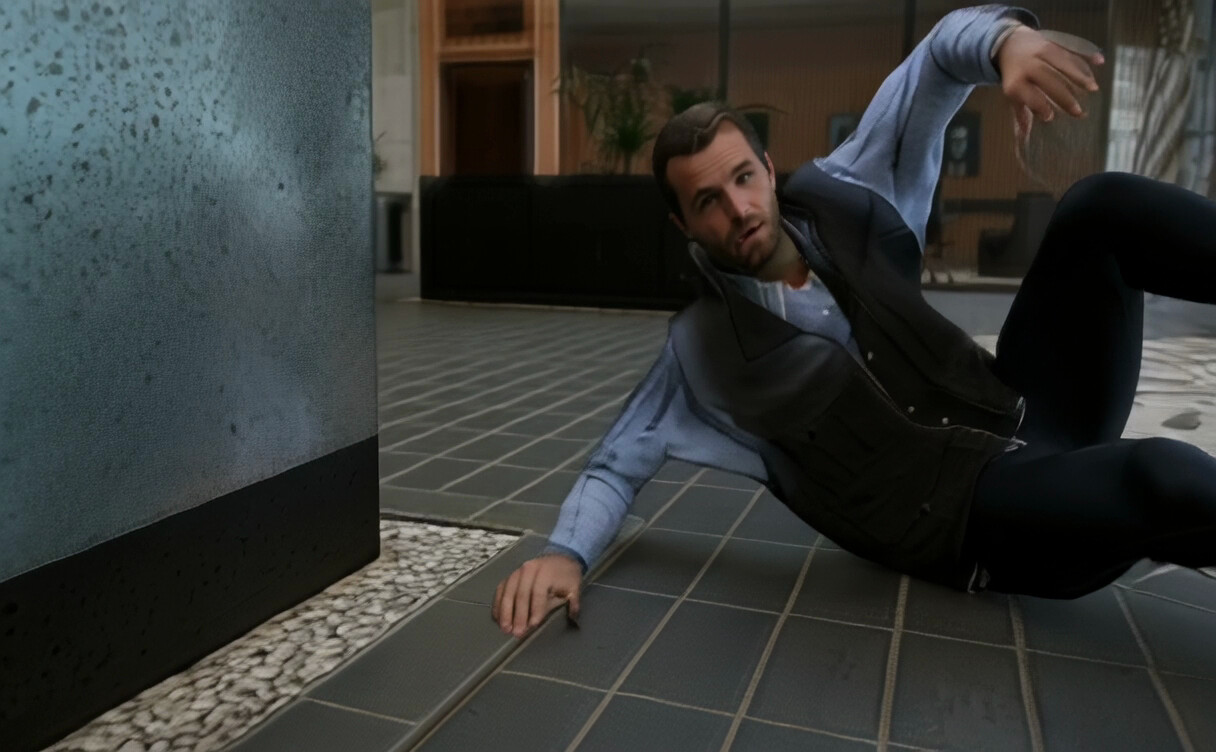} &
        \includegraphics[width=0.31\linewidth]{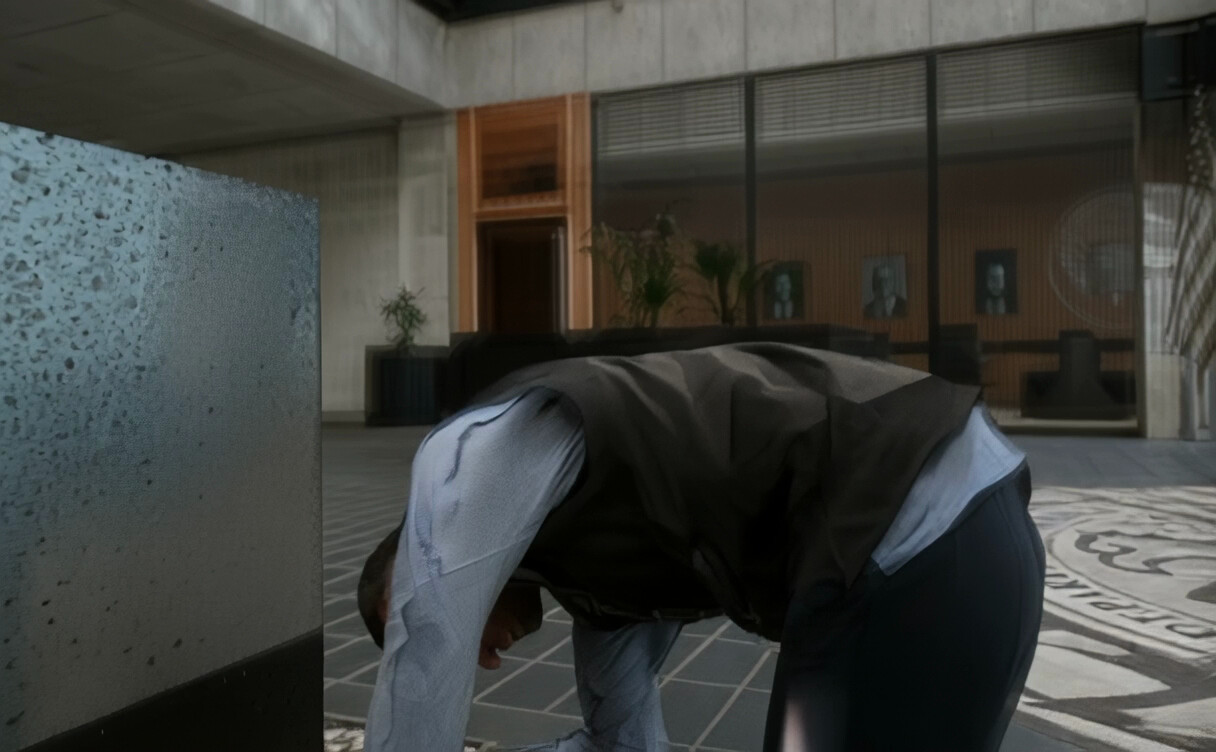} \\
        \smash{\rotatebox{90}{\small\hspace{8pt} Depth}} &
        \includegraphics[width=0.31\linewidth]{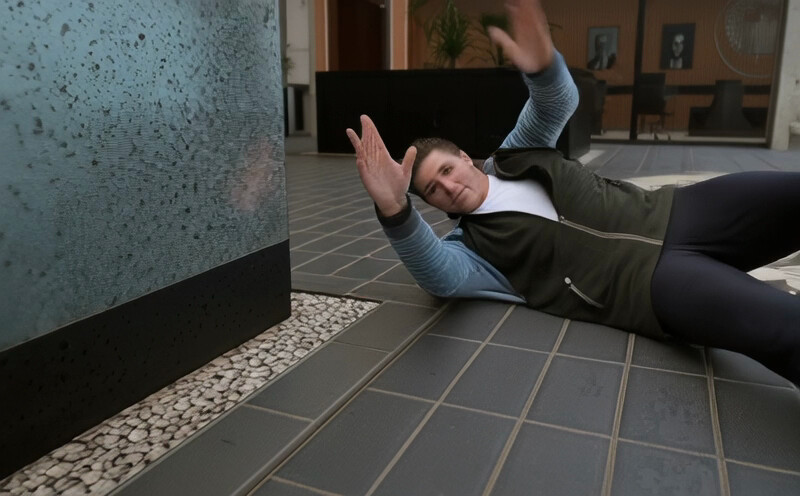} &
        \includegraphics[width=0.31\linewidth]{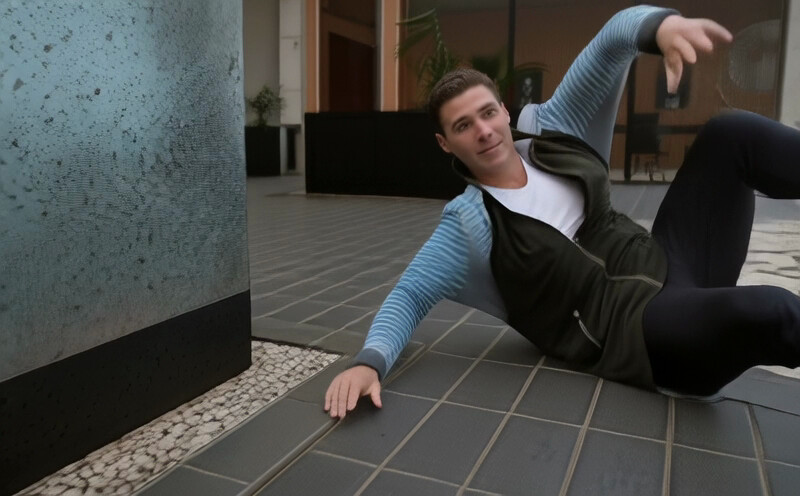} &
        \includegraphics[width=0.31\linewidth]{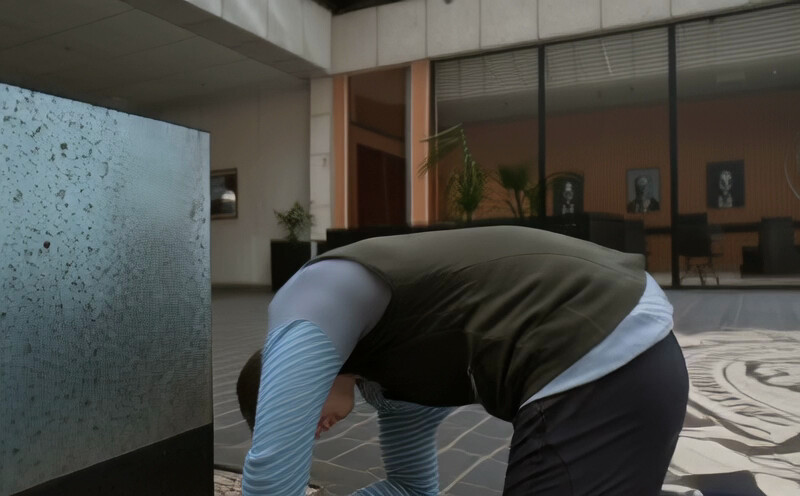} \\
        \smash{\rotatebox{90}{\small Mult. Anchors}} &
        \includegraphics[width=0.31\linewidth]{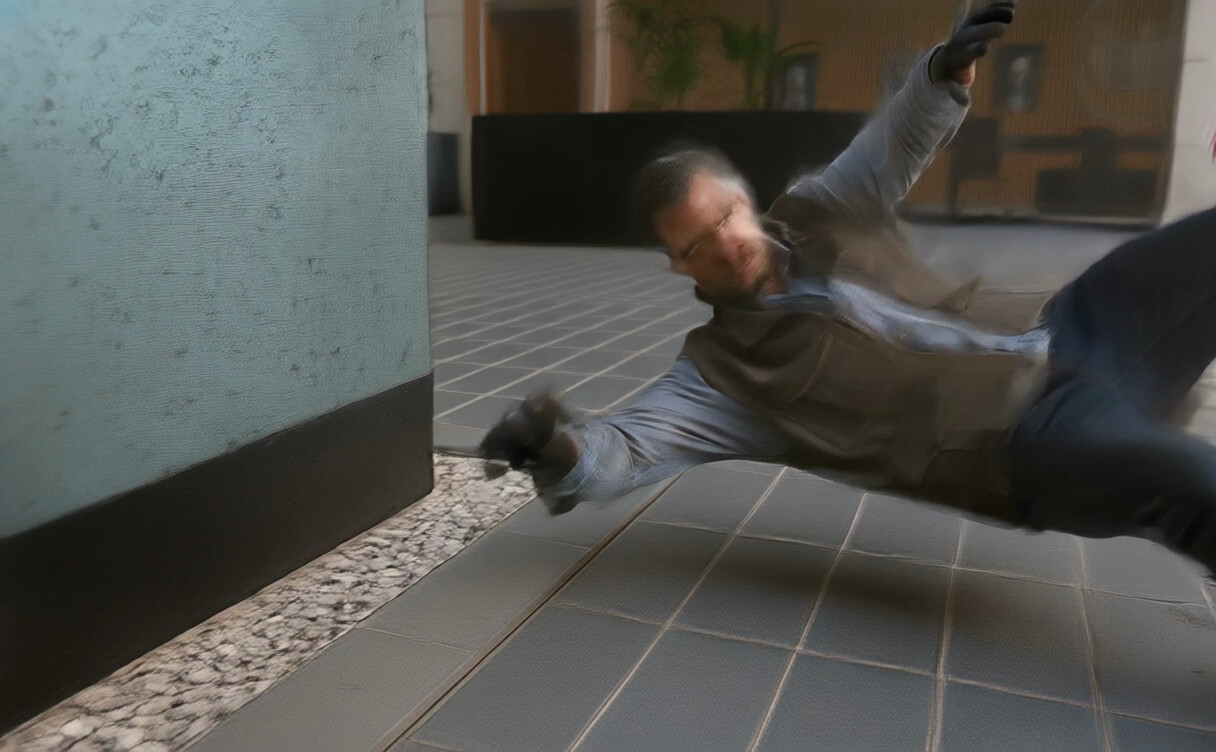} &
        \includegraphics[width=0.31\linewidth]{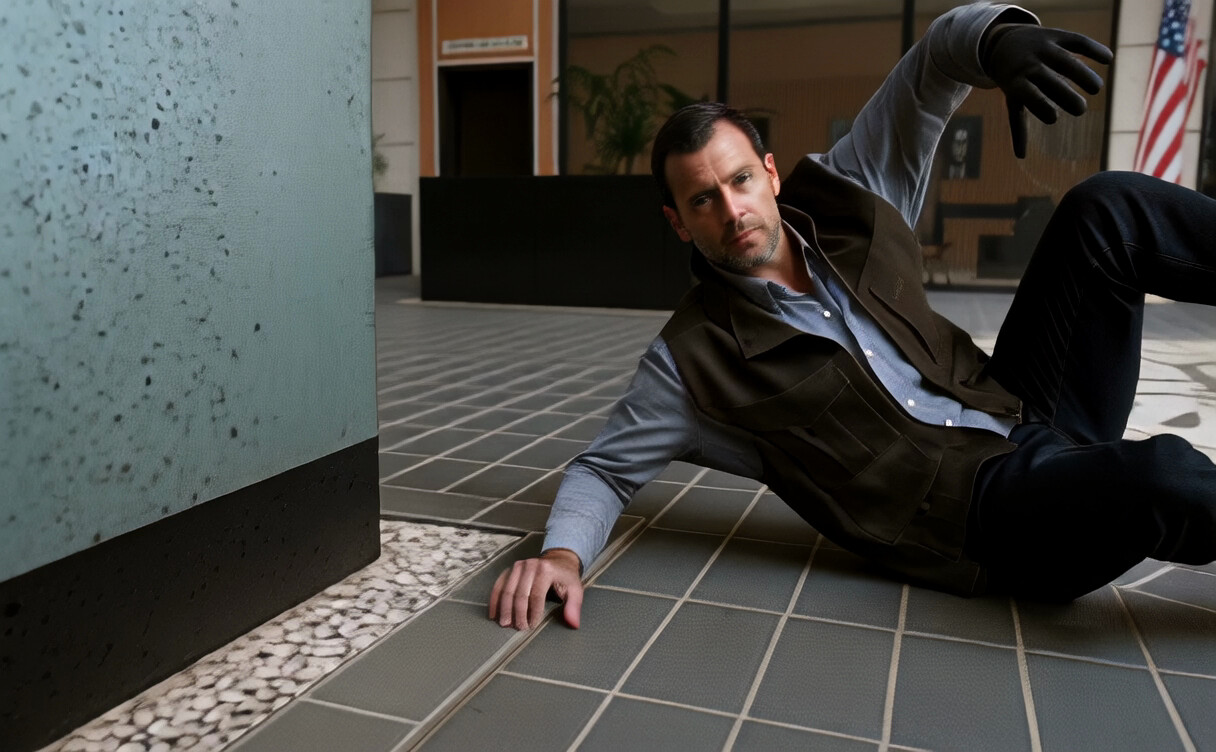} &
        \includegraphics[width=0.31\linewidth]{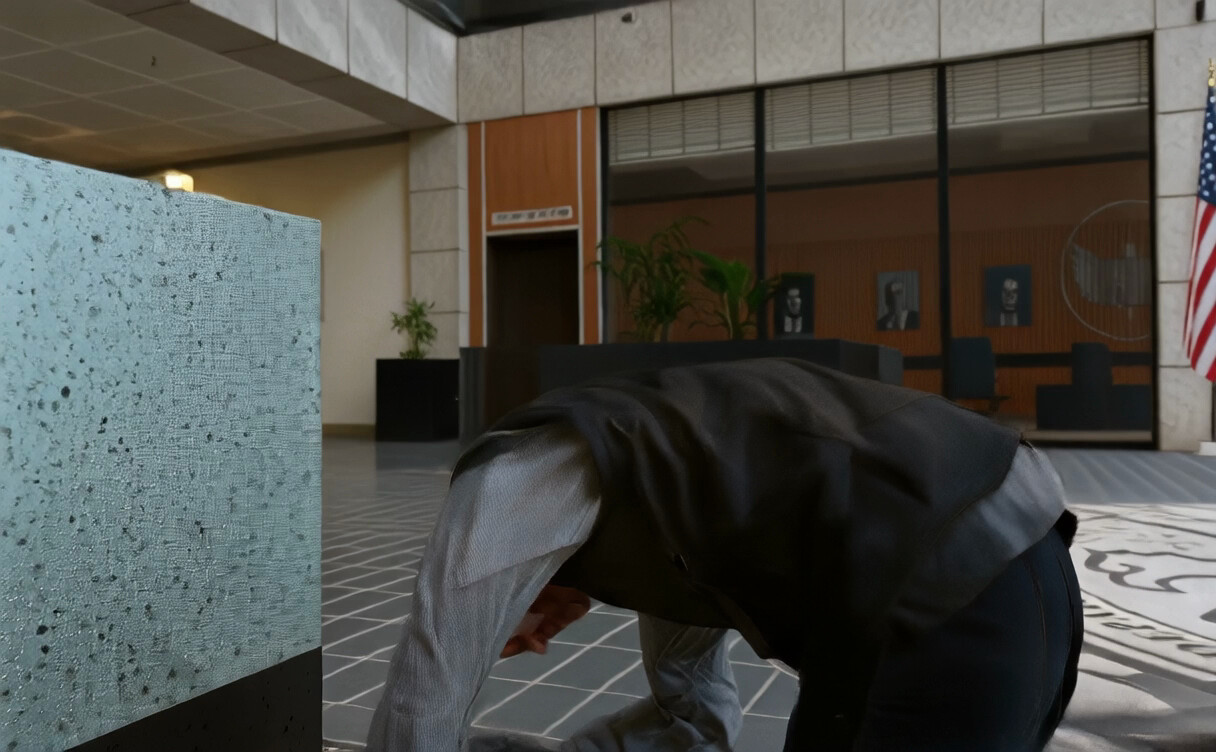} \\[4pt]
        \smash{\rotatebox{90}{\small\hspace{8pt} Source}} &
        \includegraphics[width=0.31\linewidth]{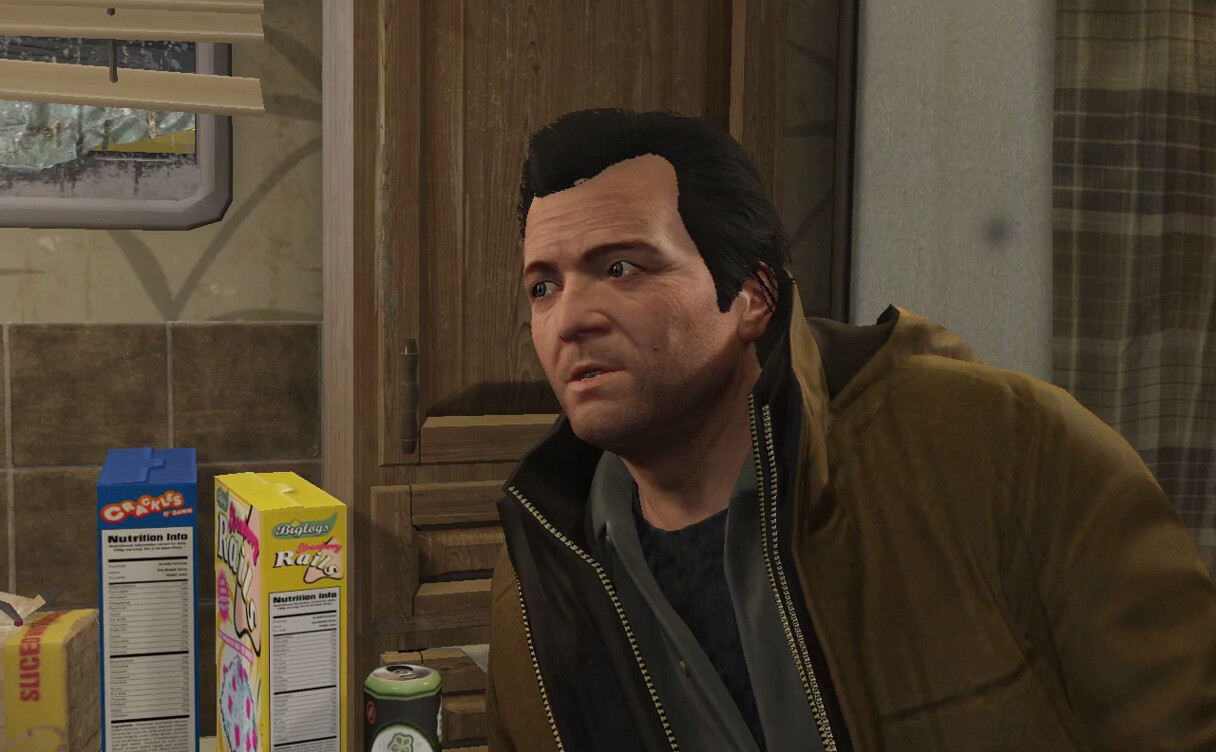} &
        \includegraphics[width=0.31\linewidth]{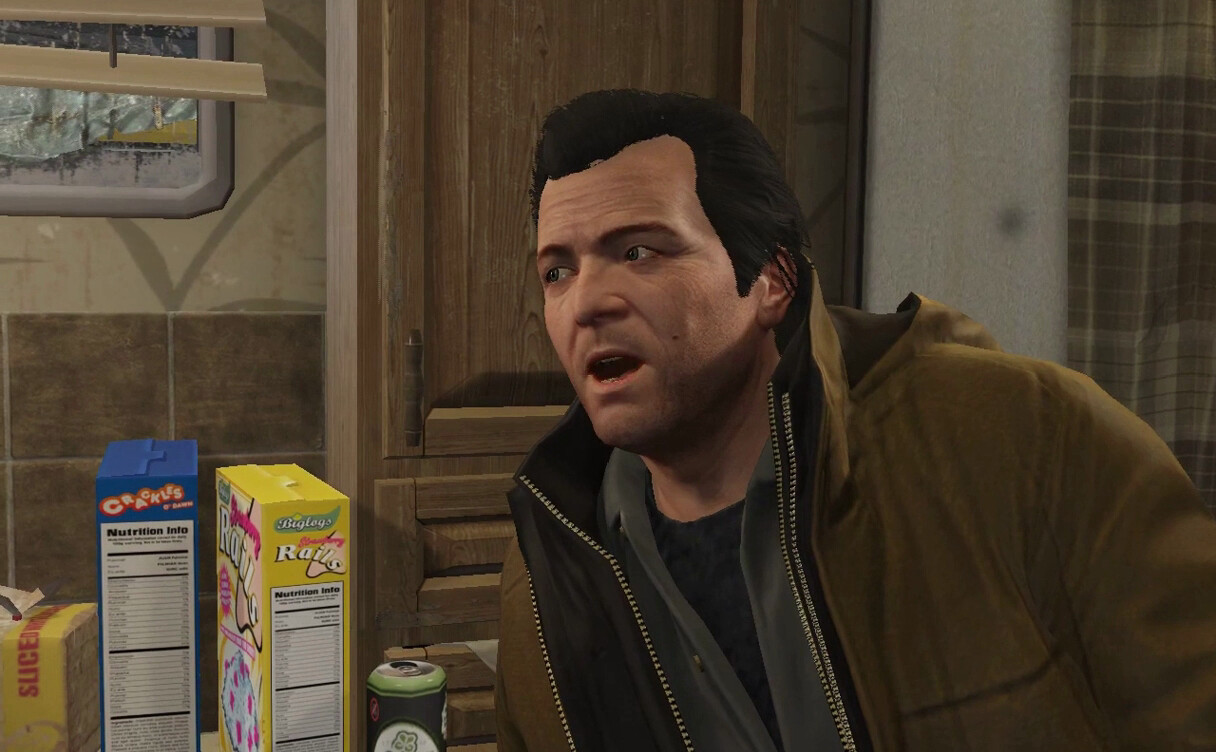} &
        \includegraphics[width=0.31\linewidth]{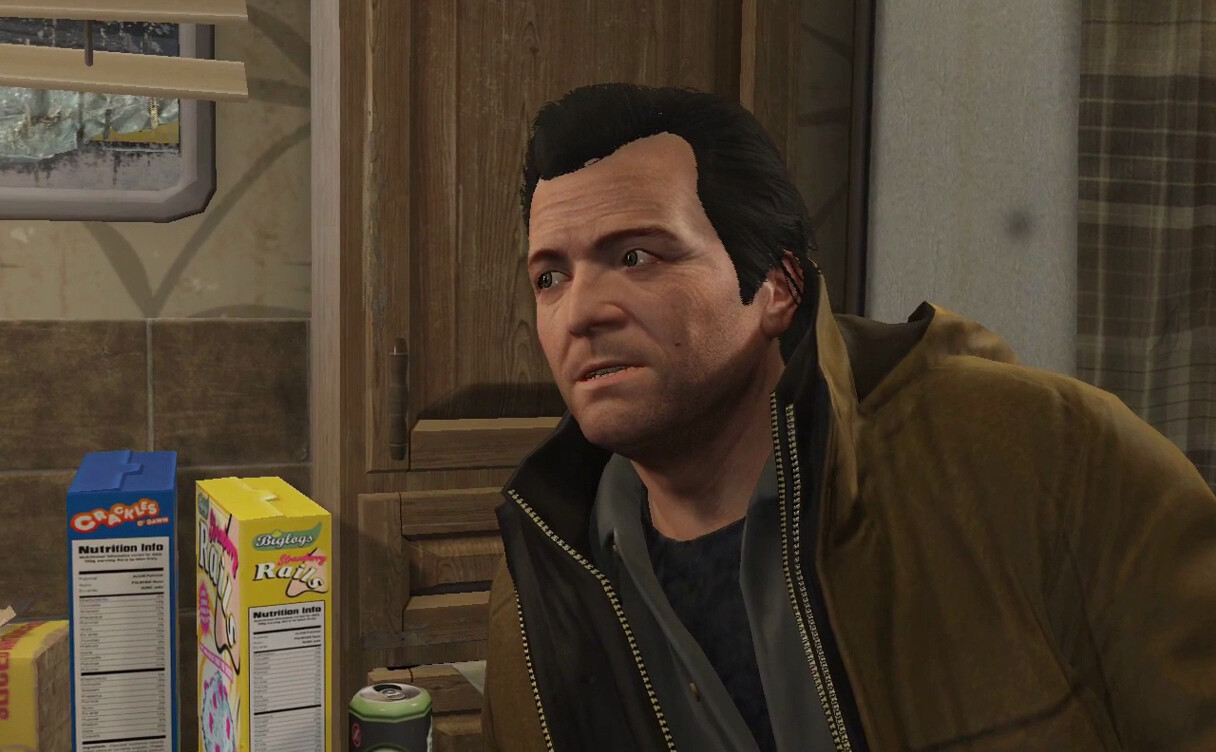} \\
        \smash{\rotatebox{90}{\small Edges (ours)}} &
        \includegraphics[width=0.31\linewidth]{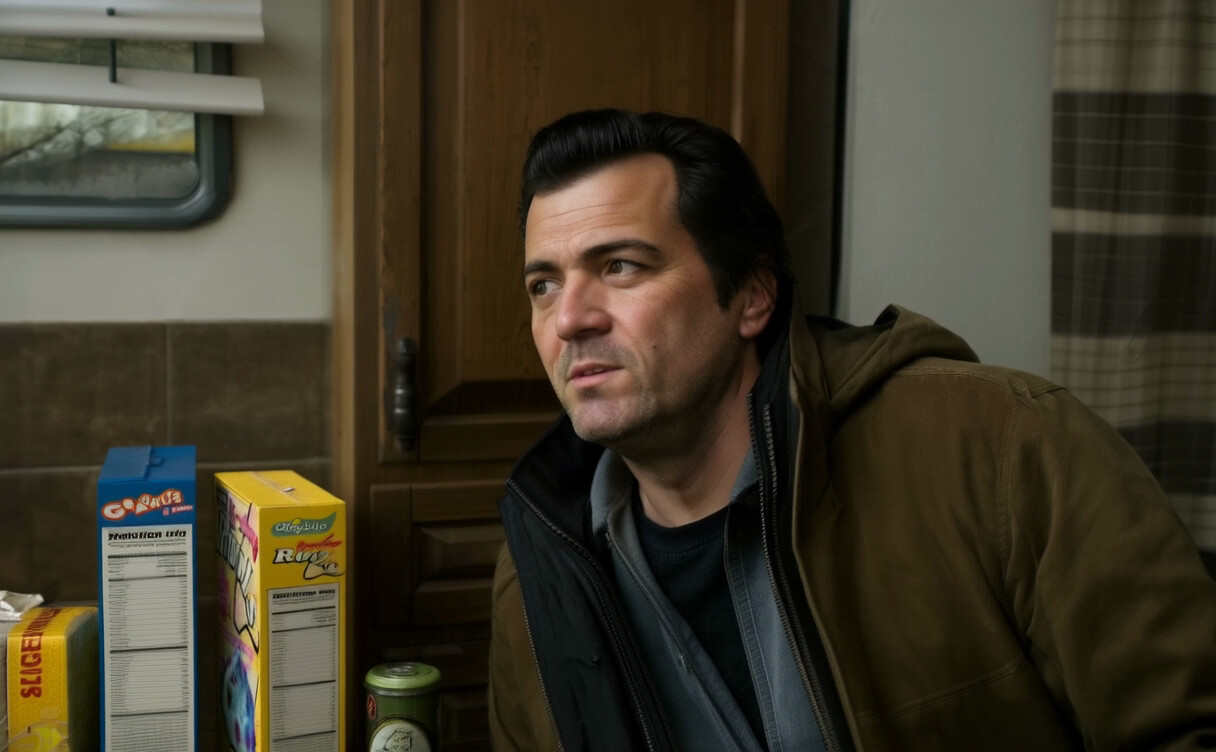} &
        \includegraphics[width=0.31\linewidth]{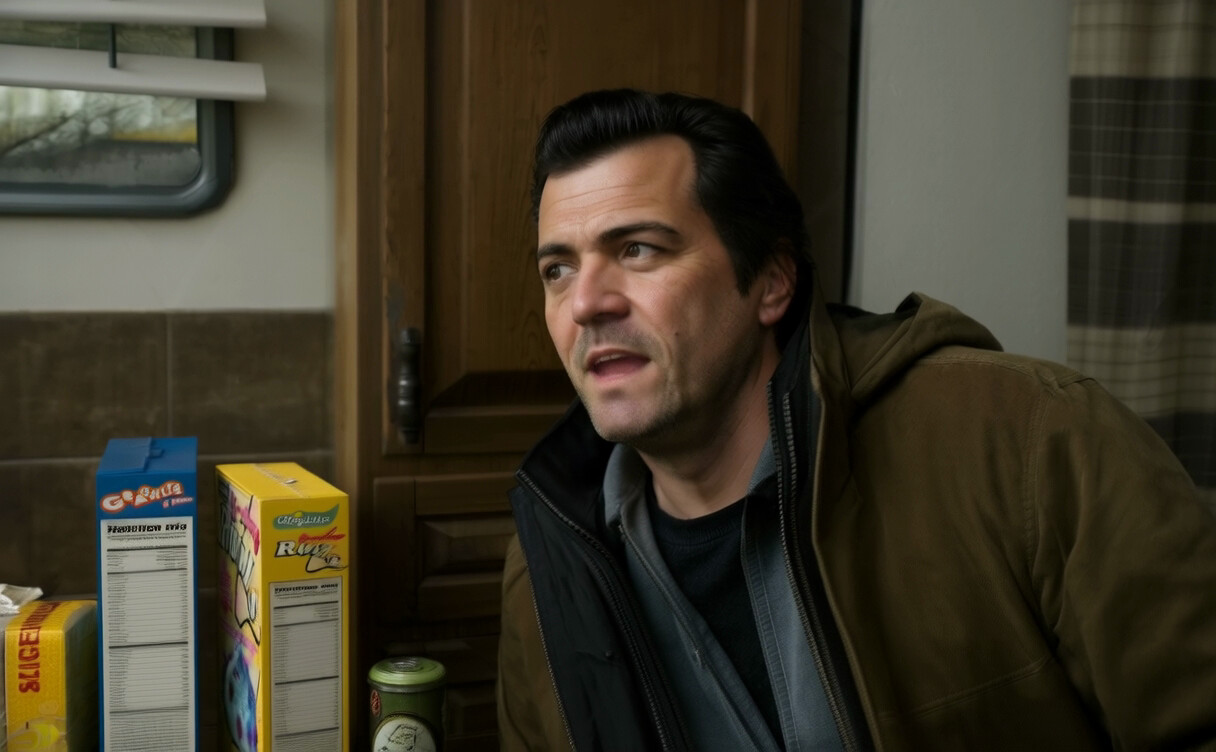} &
        \includegraphics[width=0.31\linewidth]{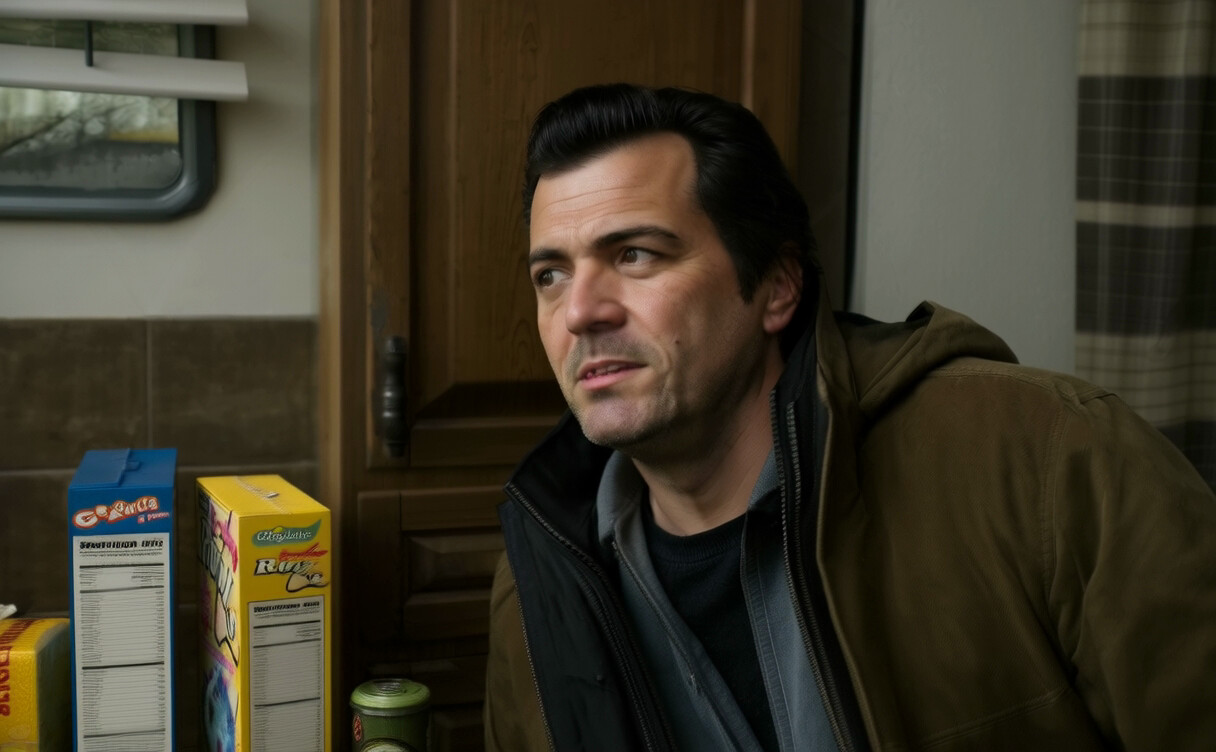} \\
        \smash{\rotatebox{90}{\small\hspace{8pt} Depth}} &
        \includegraphics[width=0.31\linewidth]{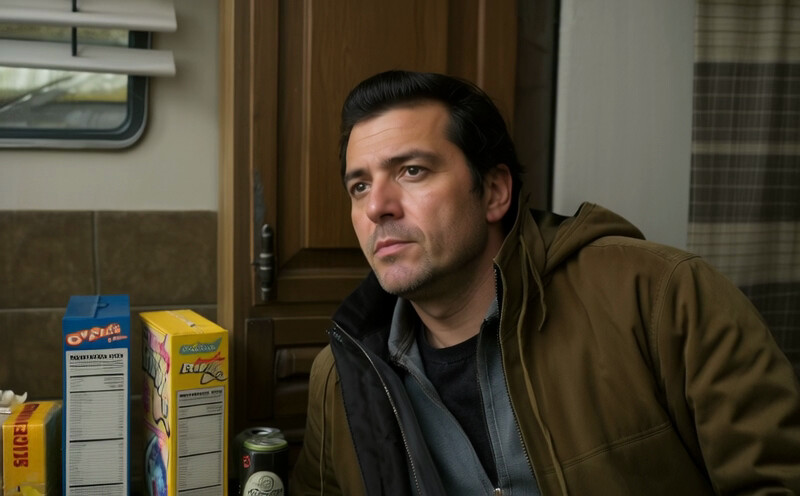} &
        \includegraphics[width=0.31\linewidth]{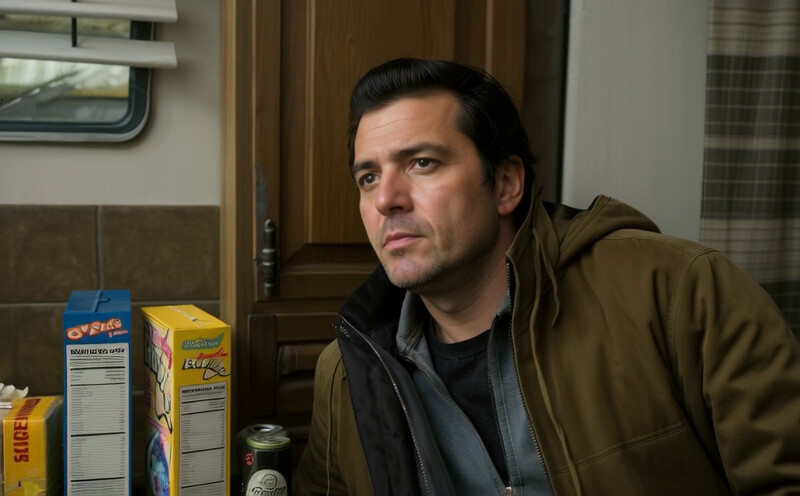} &
        \includegraphics[width=0.31\linewidth]{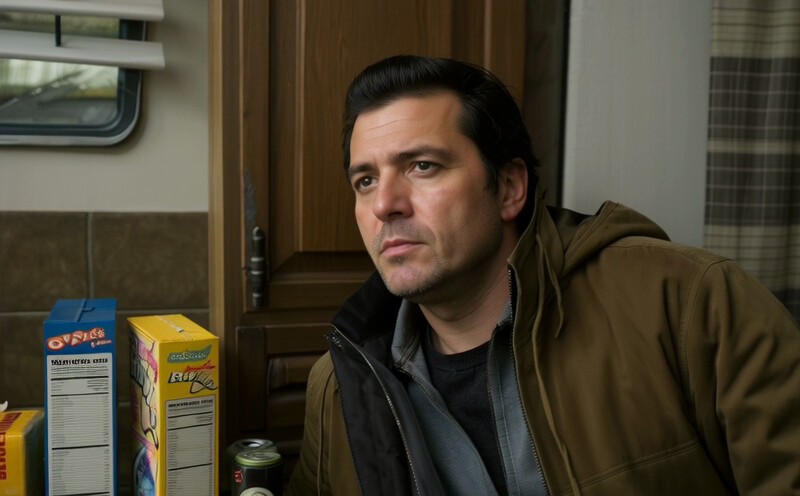} \\
        \smash{\rotatebox{90}{\small Mult. Anchors}} &
        \includegraphics[width=0.31\linewidth]{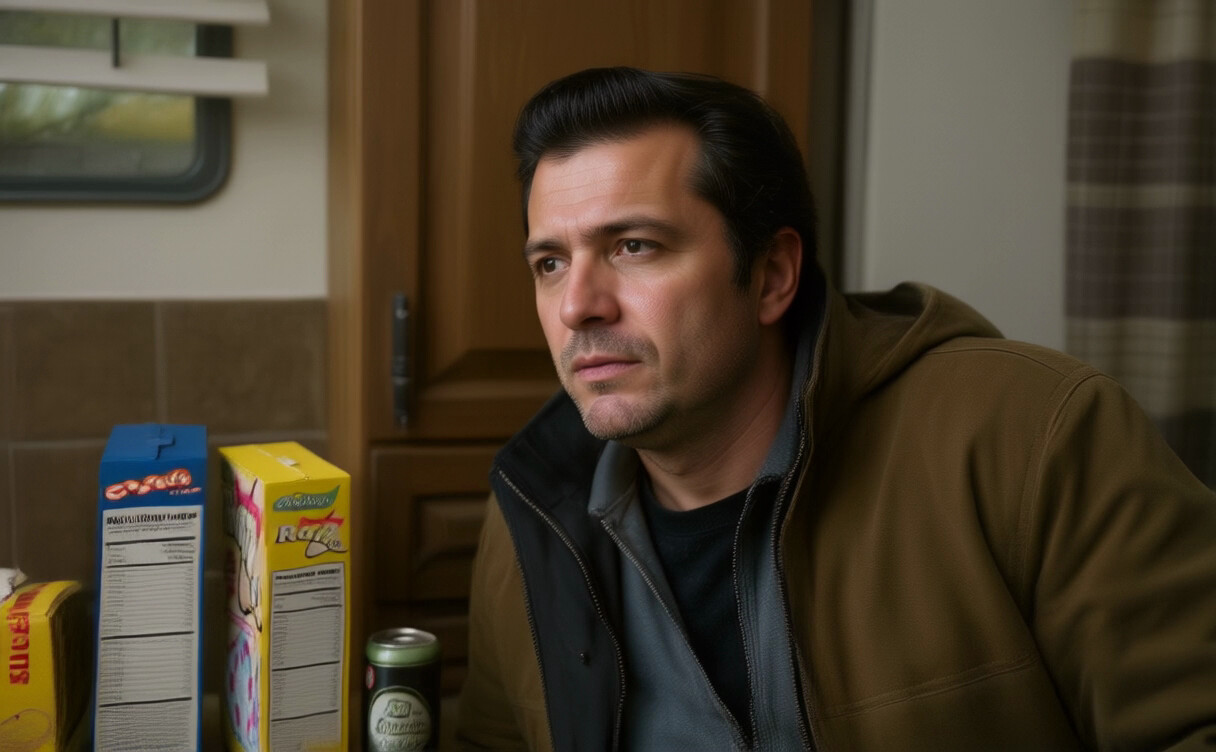} &
        \includegraphics[width=0.31\linewidth]{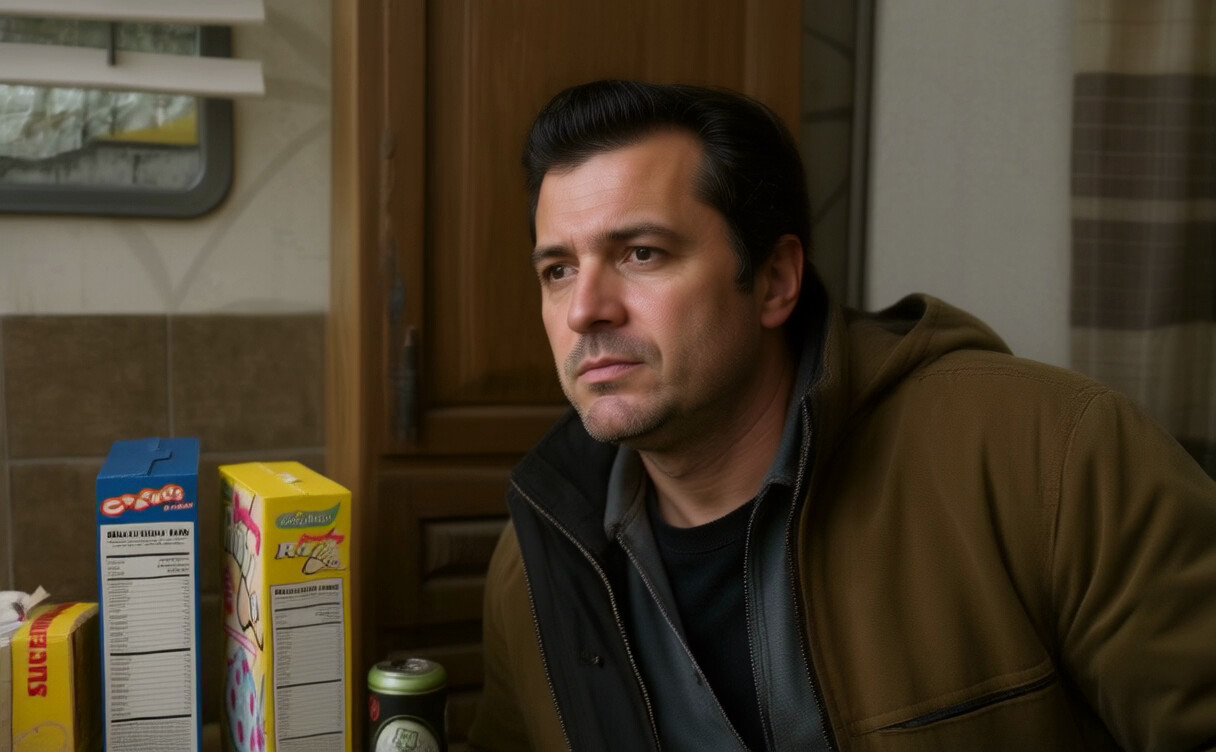} &
        \includegraphics[width=0.31\linewidth]{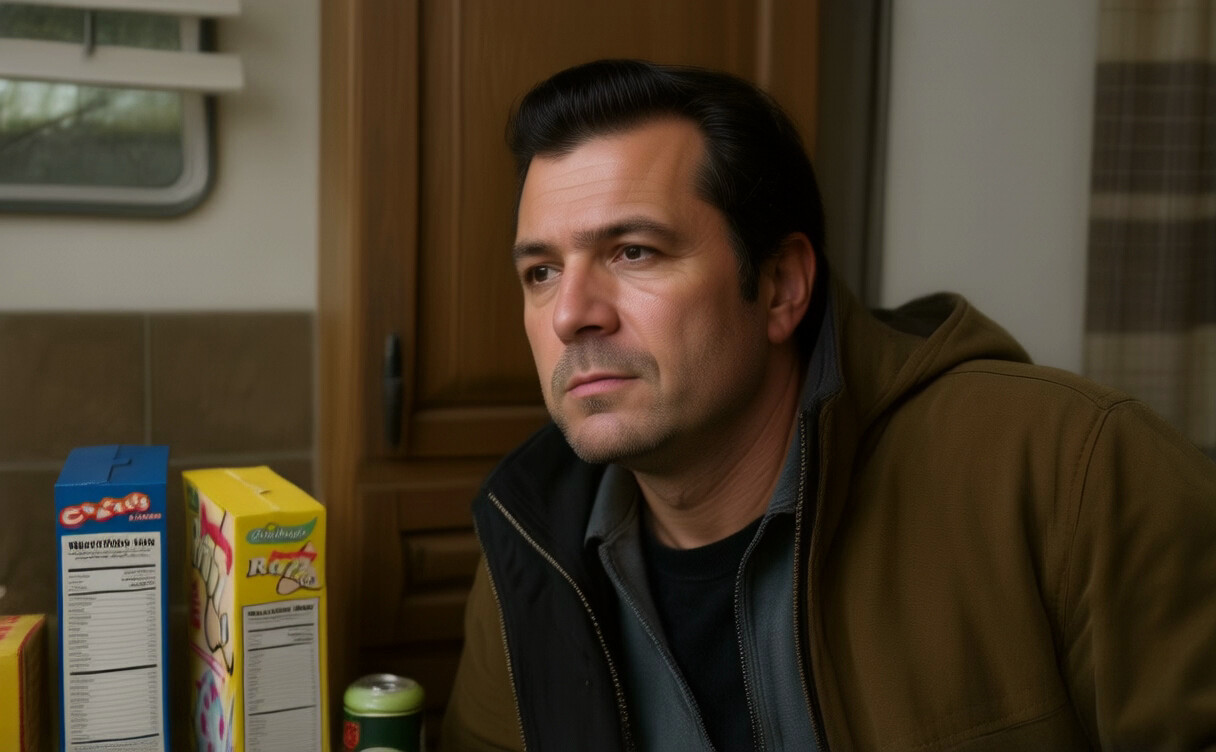} \\
    \end{tabular}
    \caption{\textbf{Data generation ablation.}
    We ablate sparse-to-dense propagation for training pair generation, comparing multiple-anchor editing, depth conditioning, and edge conditioning for VACE. Multiple-anchor editing leads to temporal flickering and fluctuations in identity. Depth conditioning loses facial expression and facial structure, often failing to preserve identity. In contrast, edge conditioning preserves facial details more reliably and produces the most stable results across the sequence.}
    \label{fig:ablation_datagen}
\end{figure}

%% file: figures/ablation_pipeline_vs_model.tex
\begin{figure}[h]
    \centering
    \setlength{\tabcolsep}{0pt}
    \renewcommand{\arraystretch}{0}
    \begin{tabular}{@{}c@{\hspace{4pt}} c@{}c@{}c@{}}
        & \small Frame 1 & \small Frame 2 & \small Frame 3 \\[2pt]
        \smash{\rotatebox{90}{\small\hspace{4pt} Source}} &
        \includegraphics[width=0.31\linewidth]{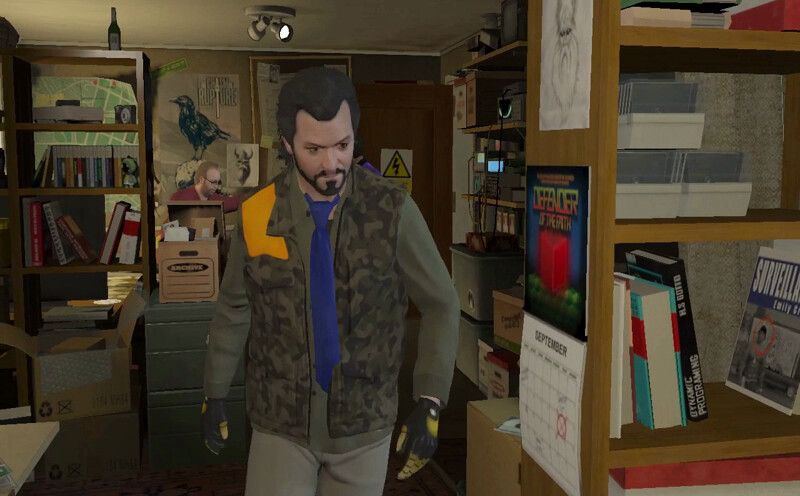} &
        \includegraphics[width=0.31\linewidth]{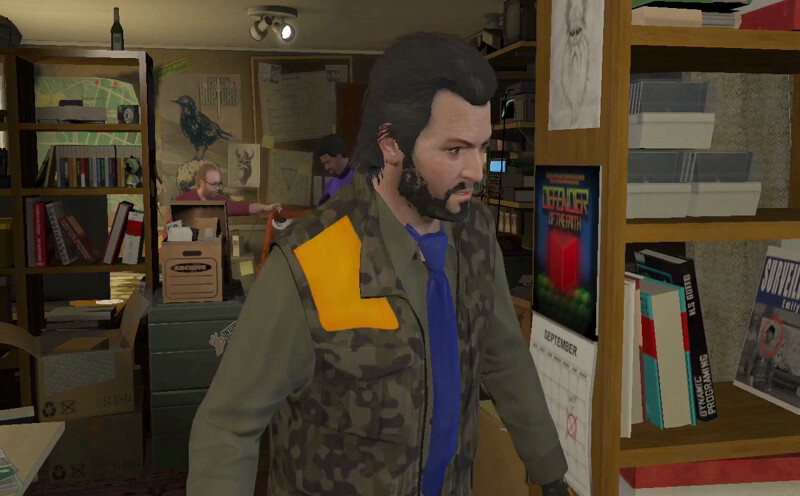} &
        \includegraphics[width=0.31\linewidth]{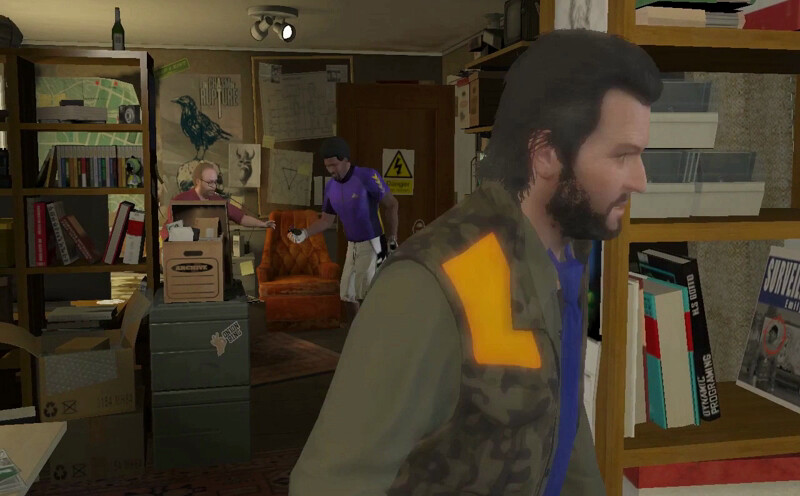} \\
        \smash{\rotatebox{90}{\small RealMaster}} &
        \includegraphics[width=0.31\linewidth]{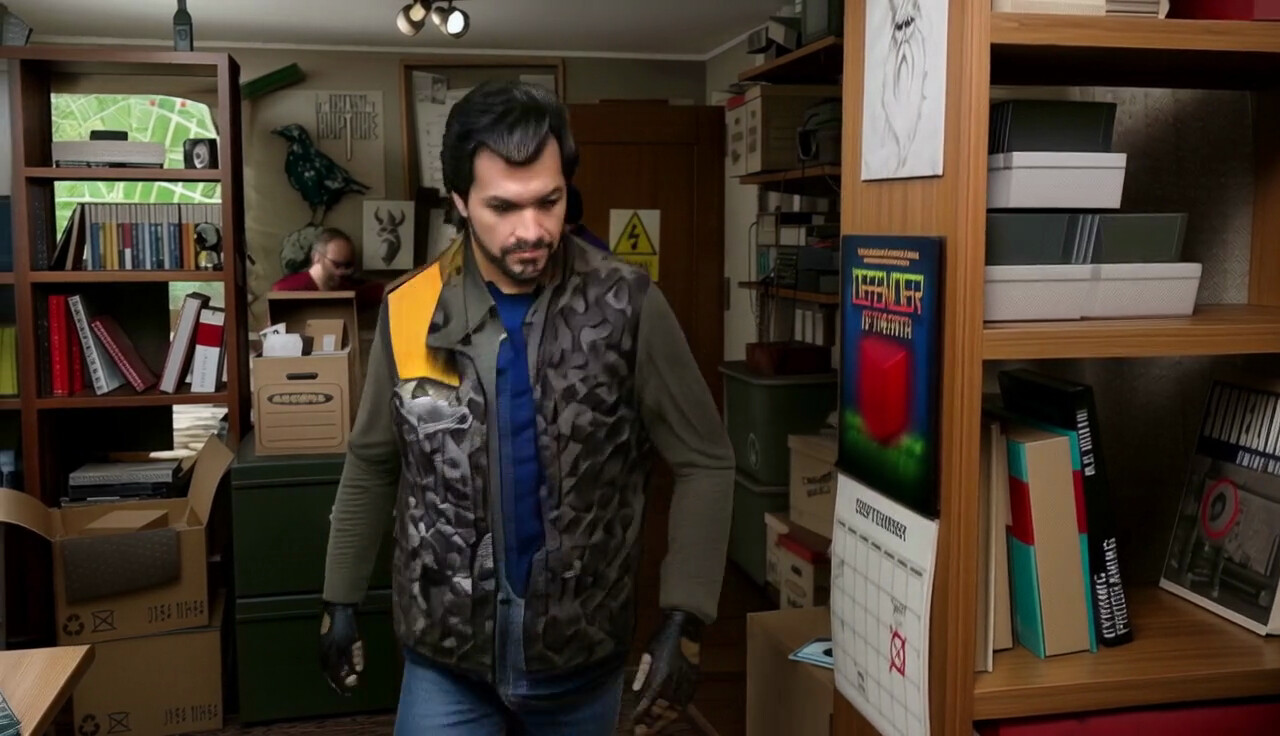} &
        \includegraphics[width=0.31\linewidth]{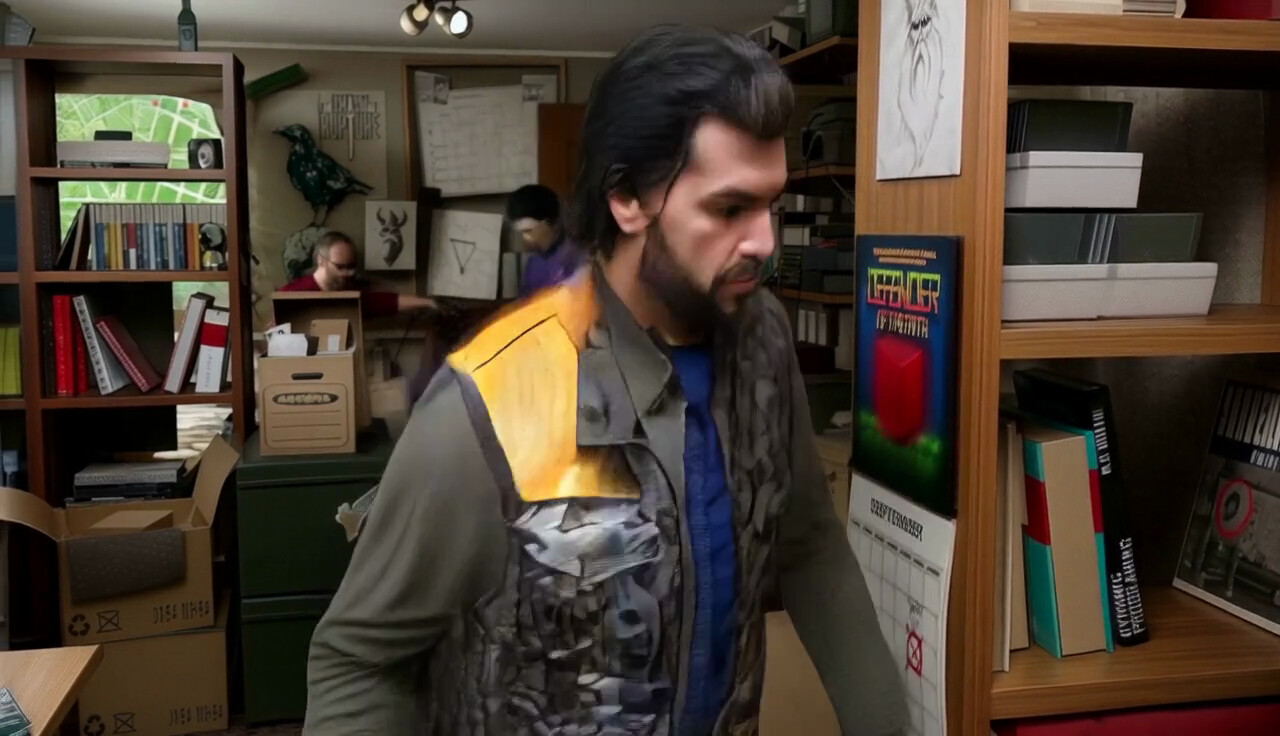} &
        \includegraphics[width=0.31\linewidth]{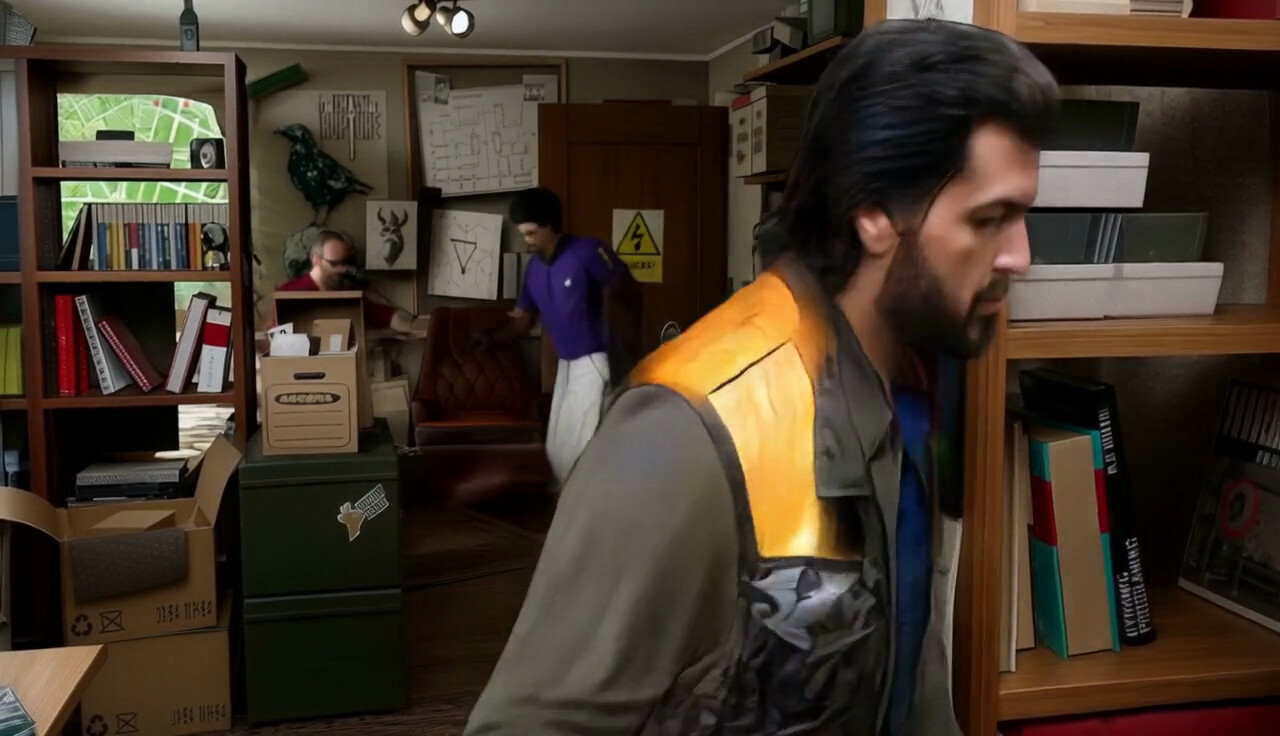} \\
        \smash{\rotatebox{90}{\small\hspace{8pt} Data}} &
        \includegraphics[width=0.31\linewidth]{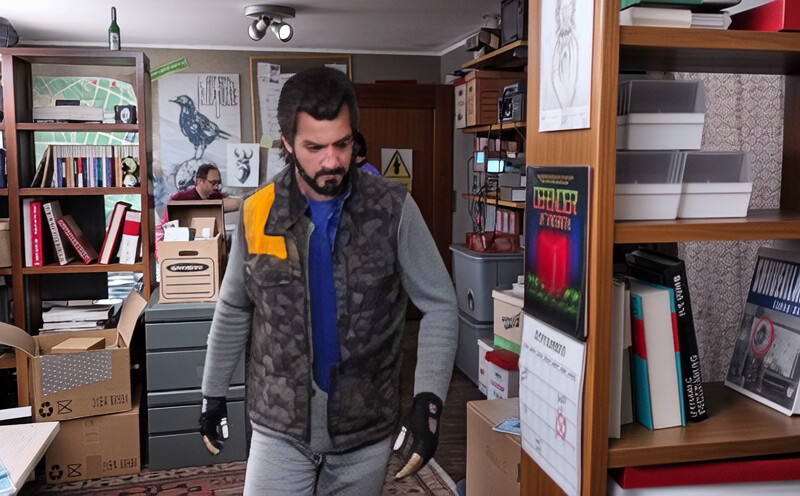} &
        \includegraphics[width=0.31\linewidth]{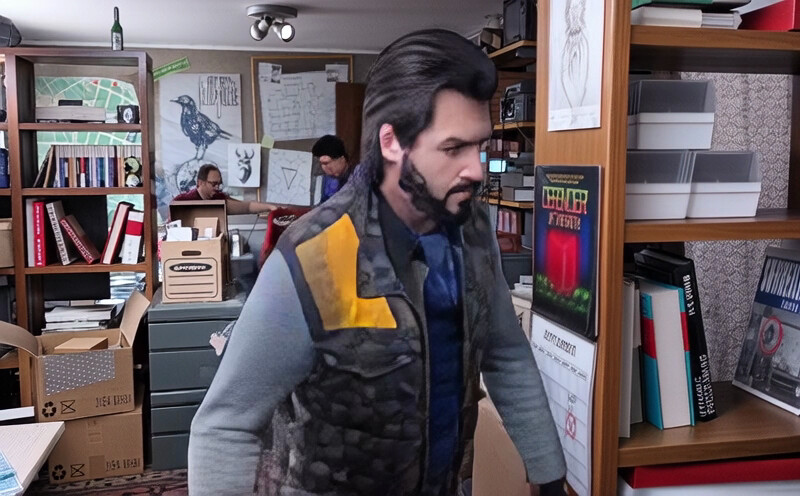} &
        \includegraphics[width=0.31\linewidth]{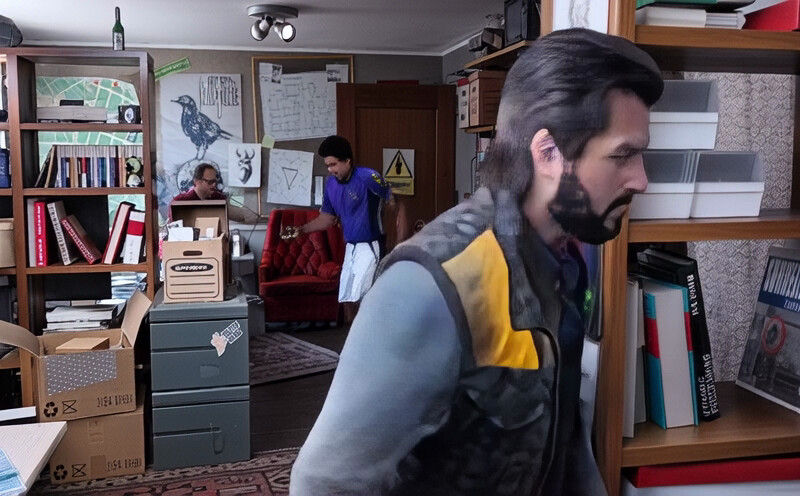} \\[4pt]
        \smash{\rotatebox{90}{\small\hspace{4pt} Source}} &
        \includegraphics[width=0.31\linewidth]{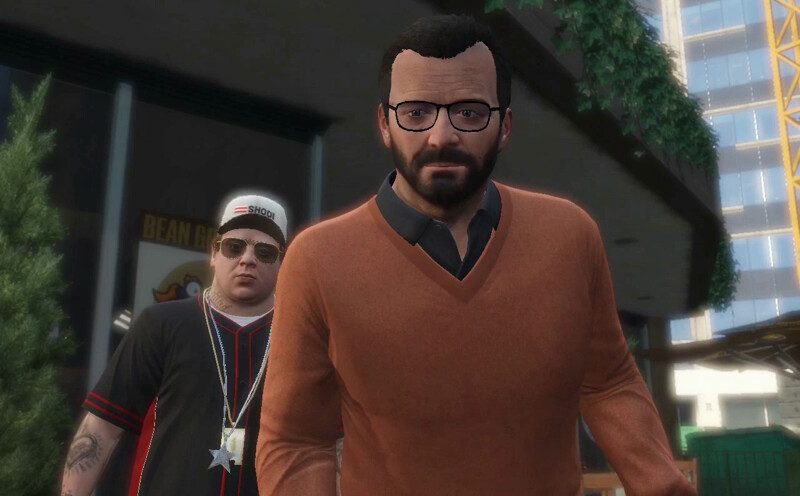} &
        \includegraphics[width=0.31\linewidth]{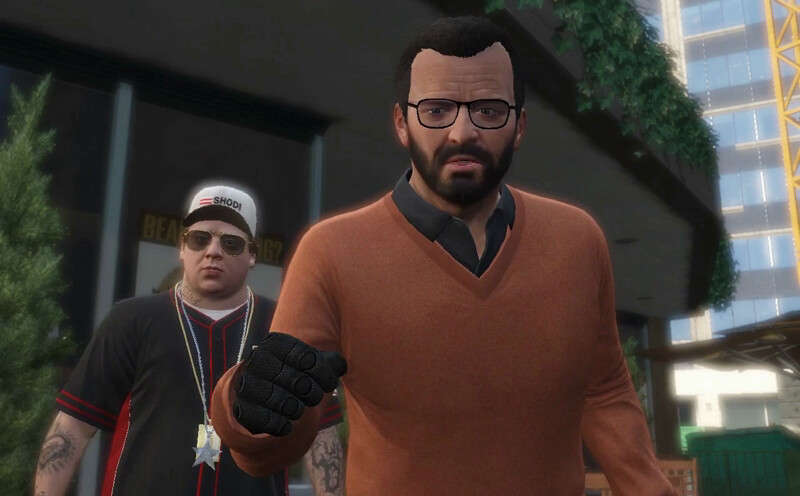} &
        \includegraphics[width=0.31\linewidth]{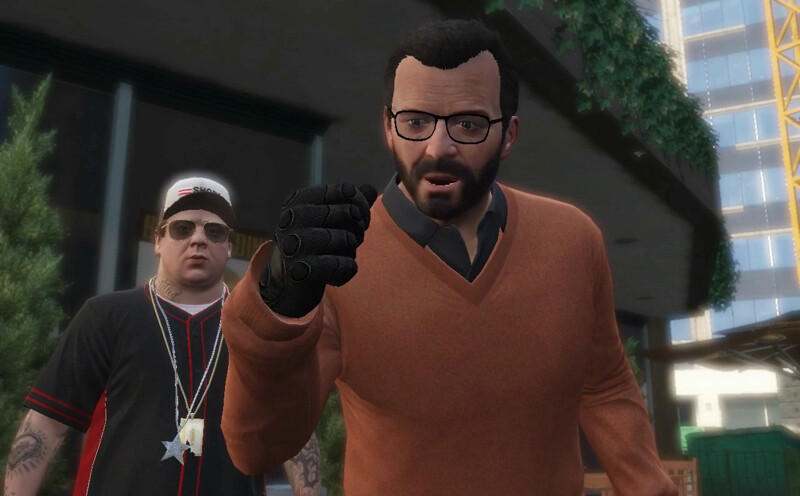} \\
        \smash{\rotatebox{90}{\small RealMaster}} &
        \includegraphics[width=0.31\linewidth]{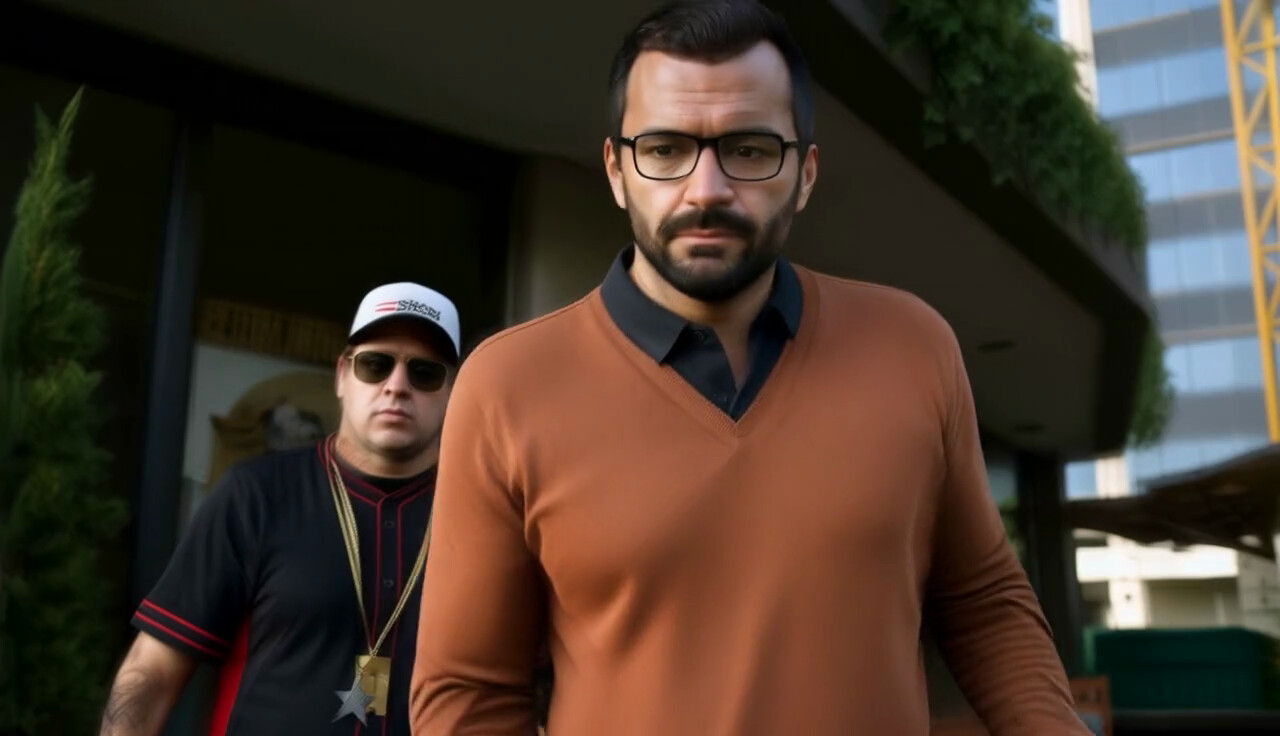} &
        \includegraphics[width=0.31\linewidth]{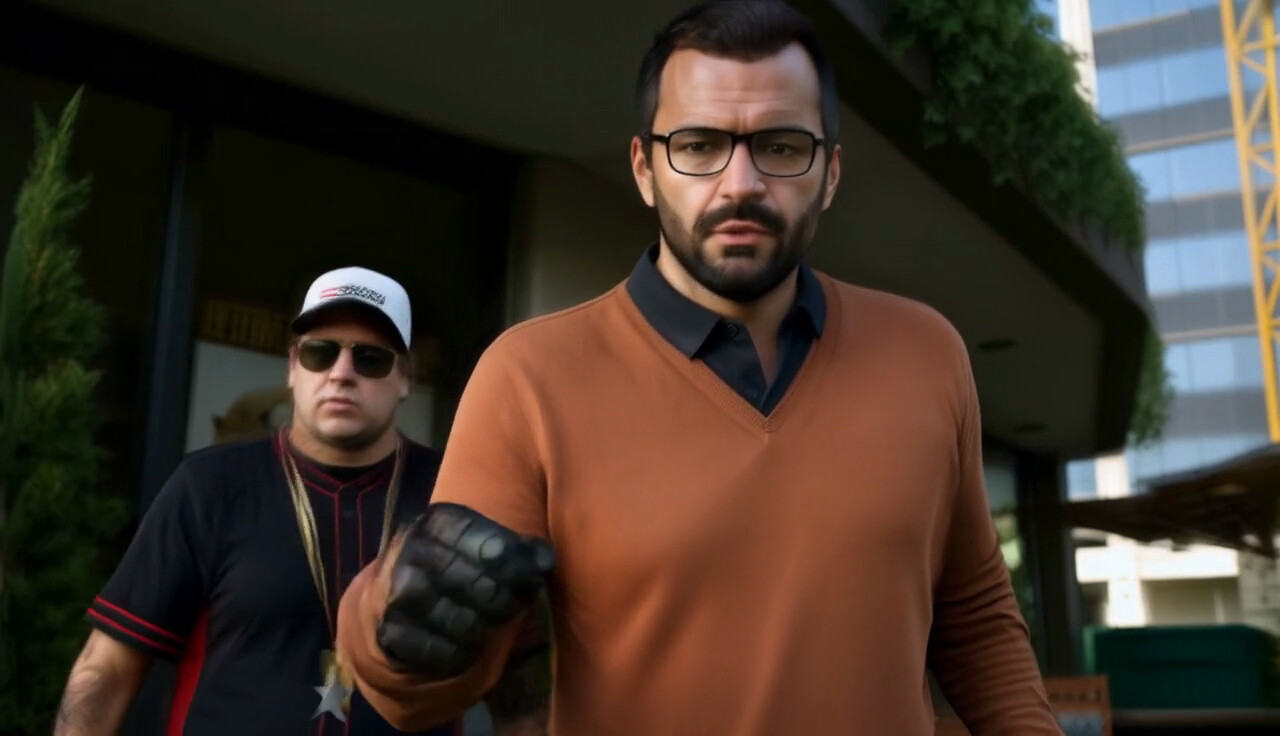} &
        \includegraphics[width=0.31\linewidth]{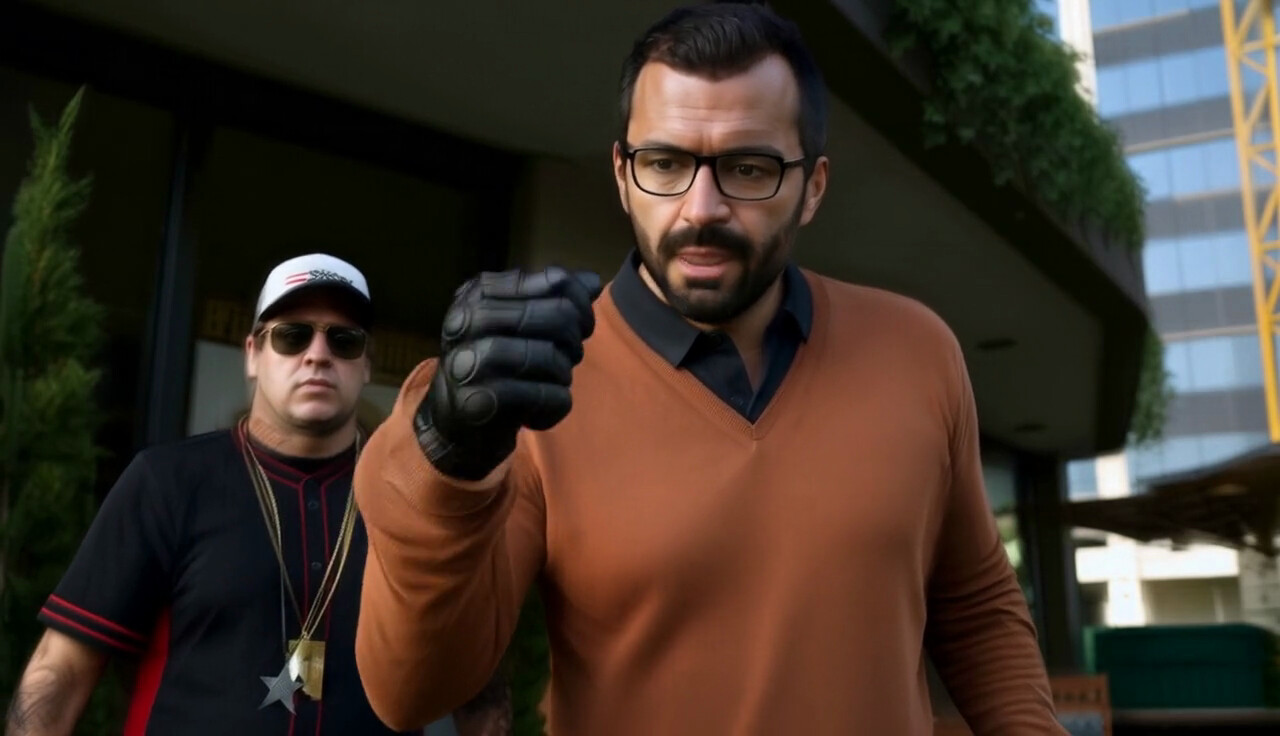} \\
        \smash{\rotatebox{90}{\small\hspace{8pt} Data}} &
        \includegraphics[width=0.31\linewidth]{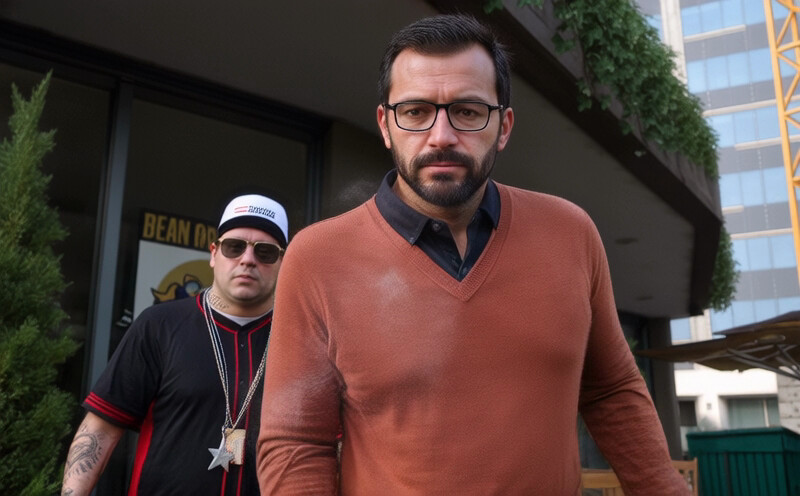} &
        \includegraphics[width=0.31\linewidth]{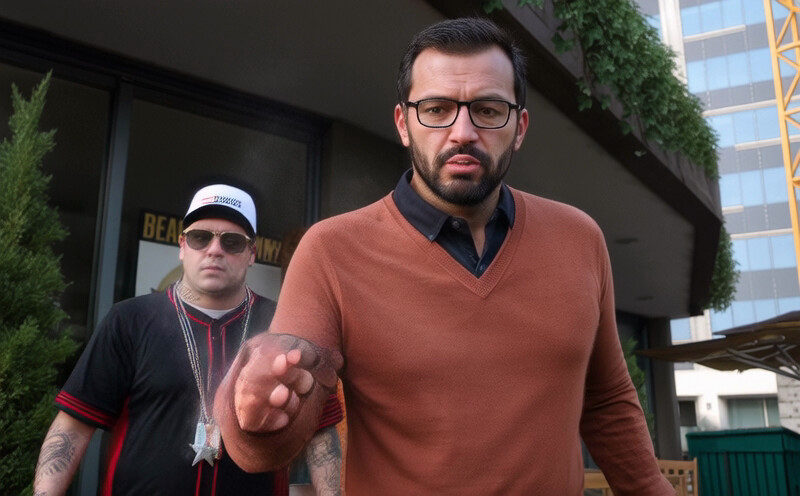} &
        \includegraphics[width=0.31\linewidth]{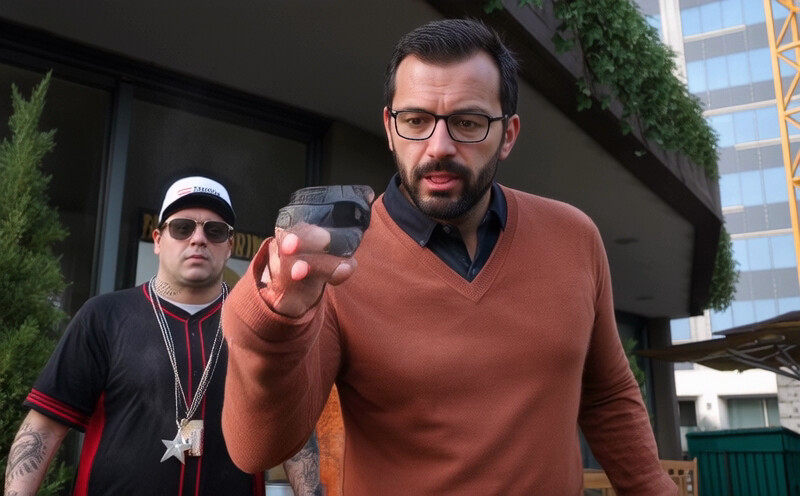} \\
    \end{tabular}
    \caption{\textbf{Model vs. data pipeline comparison.}
    We compare direct use of the data generation pipeline to inference with our trained model.
    \textbf{Top}: Our trained model produces a more faithful translation, preserving object identity, the color palette, and lighting.
    \textbf{Bottom}: The data generation pipeline fails when new objects (e.g., gloves) appear mid-sequence, since it relies only on two boundary anchors to define appearance.}
    \label{fig:ablation_pipeline_vs_model}
\end{figure}

%% file: tables/ablation_metrics.tex
\begin{table}[h]
    \centering
    \caption{\textbf{Ablation study results.} We compare multiple variants of the data generation pipeline and our trained model. The data variants include \textit{Multiple Anchors}, which introduces anchor-frame edits every 0.5 seconds instead of two boundary anchors, and \textit{Depth} and \textit{Edges}, which condition VACE on depth maps or edge maps, respectively. \textit{RealMaster} denotes the trained diffusion model learned from edge-based data. All variants are evaluated on the SAIL-VOS validation set.}
    \label{tab:ablation_metrics}
    \footnotesize
    \setlength{\tabcolsep}{3pt}
    \renewcommand{\arraystretch}{0.9}
    \resizebox{\columnwidth}{!}{%
    \begin{tabular}{lcccc}
        \toprule
        \textbf{Variant} 
        & \textbf{ArcFace}$\uparrow$ 
        & \textbf{DINO}$\downarrow$ 
        & \textbf{Temp. Flicker}$\uparrow$ 
        & \textbf{Motion Smooth.}$\uparrow$ \\
        \midrule
        Multiple anchors
        & 0.357
        & 33.983
        & 0.950 & 0.969 \\
        Depth
        & 0.334
        & 34.27
        & 0.952 & 0.954 \\
        Edges
        & 0.468
        & 32.29
        & 0.954 & 0.954 \\
        \midrule
        \textbf{RealMaster} 
        & \textbf{0.473}
        & \textbf{30.28} 
        & \textbf{0.976} & \textbf{0.973} \\
        \bottomrule
    \end{tabular}%
    }
\end{table}

%% file: 5_applications.tex
\section{Additional Applications}
\label{sec:applications}

Beyond standard sim-to-real translation, our approach enables capabilities 
that would require significant effort to achieve in traditional rendering pipelines.

\paragraph{\textbf{Dynamic Weather Effects.}}
Video diffusion models inherently capture rich priors about natural phenomena, 
including weather dynamics. By simply modifying the text prompt at inference time, 
our model can introduce dynamic weather effects such as rain or snow into rendered 
scenes. \cref{fig:weather_edit} shows an example of this capability. These effects include realistic details that are challenging to synthesize 
in 3D engines, such as wet surface reflections, falling raindrops, and snow 
accumulation. Traditional simulators require careful modeling of these phenomena, 
including particle systems, shader modifications, and environmental lighting 
adjustments. In contrast, our approach provides these capabilities through the 
learned priors of the video model, without any additional engineering effort.
\input{figures/weather_editing}

\paragraph{\textbf{Cross-Simulator Generalization.}}
Although our model is trained exclusively on data from SAIL-VOS, 
where the underlying engine is GTA-V, it generalizes to rendered videos from 
other simulators. We demonstrate this by applying the same trained model to scenes 
from the CARLA-LOC dataset~\citep{han2024carlalocsyntheticslamdataset}, which is
collected in the CARLA driving
simulator~\citep{dosovitskiy2017carlaopenurbandriving} and has significantly
different characteristics. Unlike SAIL-VOS, which features third-person views of
people, videos in CARLA-LOC are captured from an egocentric driving perspective
and focus on vehicles rather than pedestrians. CARLA uses a different rendering
engine with its own lighting and material models.
As shown in \cref{fig:carla}, our model successfully transforms these
scenes into photorealistic video while preserving the original structure and
dynamics, despite never seeing CARLA data during training.
This cross-simulator generalization suggests that the model learns a general
mapping from rendered to real appearance, rather than overfitting to the
specific visual characteristics of the training domain.

\input{figures/carla_generalization}

%% file: figures/weather_editing.tex
\begin{figure}[t]
  \centering
  \includegraphics[width=1\linewidth]{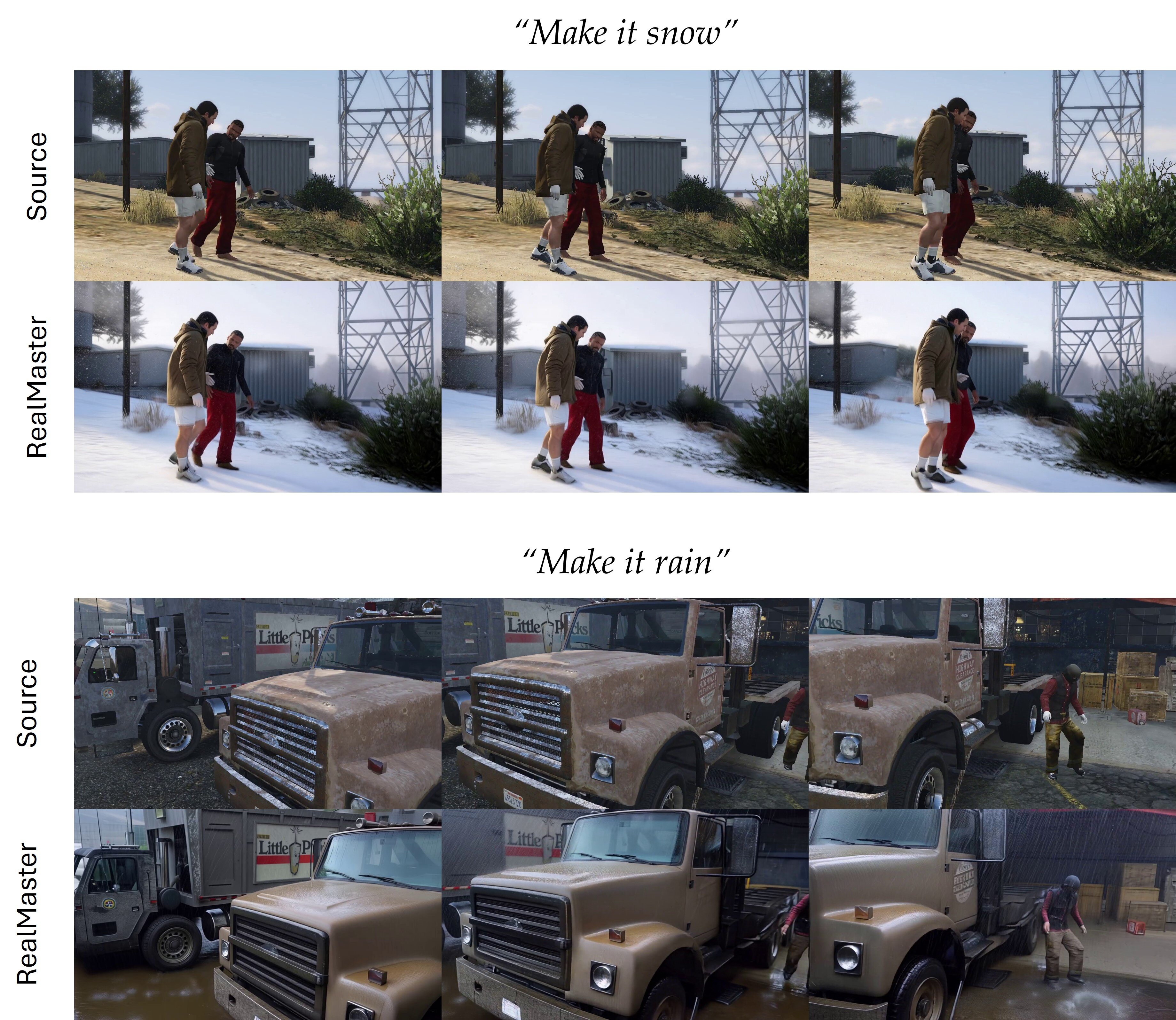}
  \caption{\textbf{Adding Weather Effects.} RealMaster can add weather effects to a given scene by changing the textual prompt, despite not being trained for this capability. The model synthesizes dynamic phenomena such as falling rain droplets and snow accumulation.}
  \label{fig:weather_edit}
\end{figure}

%% file: figures/carla_generalization.tex
\begin{figure}[t]
  \centering
  \includegraphics[width=1\linewidth]{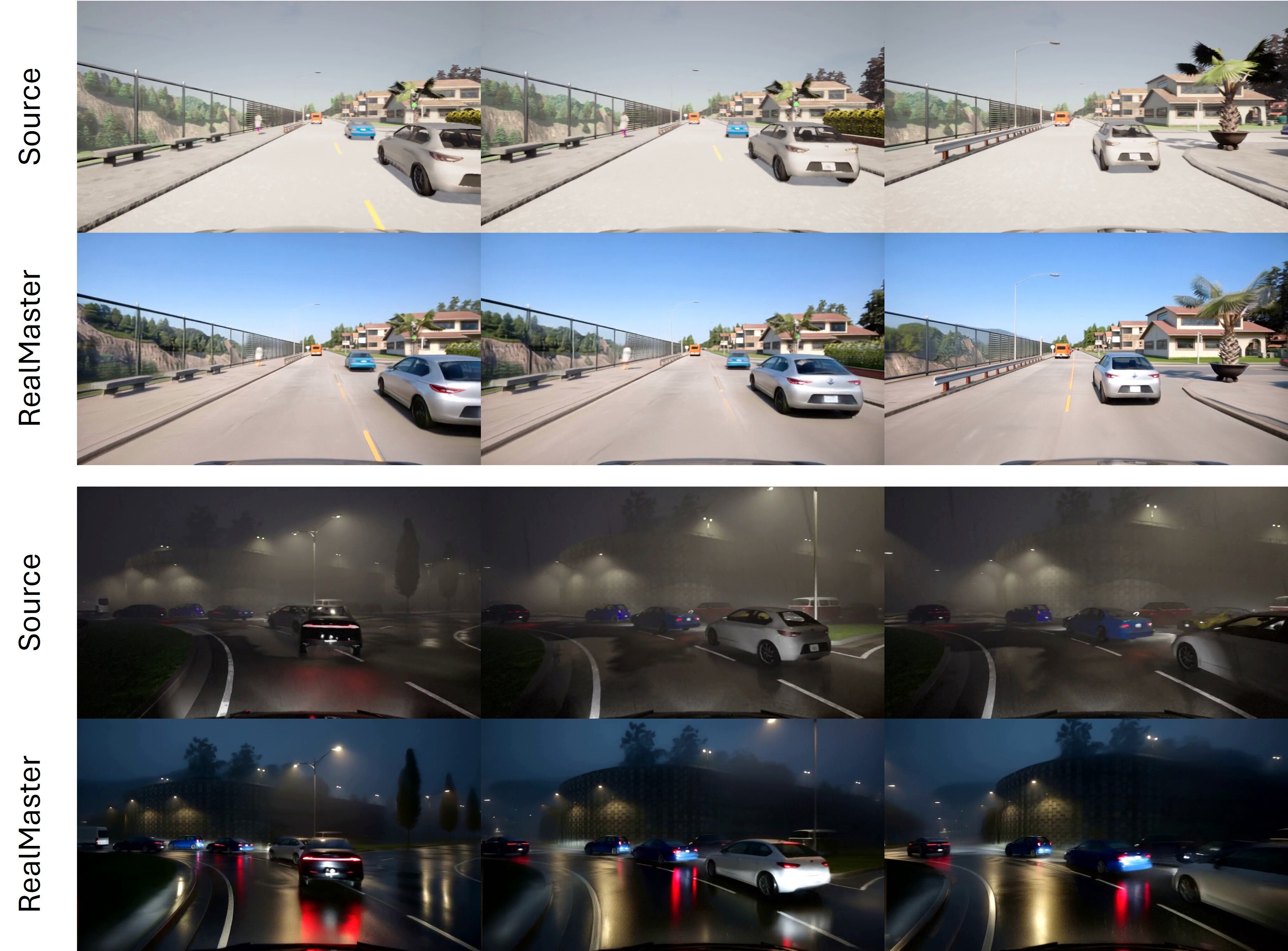}
  \caption{\textbf{Generalization to a new dataset}. We apply RealMaster, trained on SAIL-VOS, directly to CARLA videos without additional training. The latter uses a different rendering engine and features egocentric driving scenes with vehicles, in contrast to the third-person, character-centric scenes in the former. As can be seen, the model generalizes well to this different setting.}
  \label{fig:carla}
\end{figure}

%% file: 6_conclusions.tex
\section{Discussion, Limitations and Future Work}

We have presented a framework for sim-to-real video translation that lifts rendered scenes into photorealistic video while preserving underlying scene structure and dynamics. Our work is grounded in the view that sim-to-real is not merely an instance of video editing or stylization, but a problem defined by the need to reconcile two competing objectives: exact structural fidelity and global photorealistic transformation. Seen through this lens, the limitations of existing approaches stem not from incidental design choices, but from an inherent imbalance between these goals. By treating generative video models not as free-form generators, but as learned second-stage renderers operating atop explicit 3D engines, our framework separates structural control from visual realization, enabling the injection of rich real-world appearance priors without sacrificing the determinism and editability that motivate graphics pipelines.

More broadly, our results suggest that realism in generated video is not solely a matter of appearance, but of consistency maintained over time. Preserving identity, materials, and fine scale details across frames proves as critical as improving texture or lighting. Achieving such consistency requires explicit inductive bias that anchors generation to the underlying rendered structure, rather than relying on implicit regularization. Our findings also highlight the importance of data construction, where rendered structure constrains paired supervision to ensure realism is learned without compromising geometric or temporal fidelity.

Despite these advances, our approach has several limitations. First, the realism of the output is ultimately bounded by the capabilities of current image editing models, which provide photorealistic anchors during data construction; as a result, the output may still fall short of full photorealism.. In addition, while our method preserves motion and dynamics present in the rendered input, it does not explicitly reason about motion itself. In particular, complex human body locomotion, articulated gestures, and fine grained pose dynamics are inherited from the simulator rather than modeled or refined by our approach, which may limit realism in scenarios where the underlying animation is itself implausible.

Several research directions appear promising. A real-time streaming variant could enable causal sim-to-real translation with low latency, supporting interactive applications. Another direction is to move beyond appearance and address motion realism more directly. Incorporating learned priors over body dynamics and gestures could help correct rigid or synthetic motion, further narrowing the gap between simulated and real-world video.
% \section{Old Conclusions}
% \label{sec:conclusions}

% We presented RealMaster, a method for transforming synthetic 3D engine video streams into photorealistic outputs while preserving the underlying scene structure, motion, and character identity. Our approach tackles sim-to-real video translation through two key contributions: (1) a sparse-to-dense propagation strategy that constructs high-quality paired training data directly from engine-generated sequences, and (2) a LoRA-based adaptation that trains a pre-trained text-to-video diffusion model to learn a robust mapping from synthetic to photorealistic domains.

% Across extensive experiments, we show that RealMaster consistently outperforms state-of-the-art video editing and sim-to-real baselines in photorealism while better maintaining identity consistency and input structure. Moreover, the trained model generalizes beyond the limitations of the data generation pipeline, handling objects that appear mid-sequence and enabling inference without access to future keyframes.
% RealMaster focuses on appearance-level sim-to-real translation, while it
% preserves the motion and dynamics of the input video. The realism of the
% output is bounded by the capabilities of current image editing models.
% Looking ahead, a real-time streaming variant would enable causal sim-to-real
% translation with low latency for interactive applications. Extending beyond
% appearance to improve motion realism is another direction, which would
% address cases where rendered dynamics appear rigid or synthetic.

%% file: figures/figure_page_1.tex
\begin{figure*}[t]
    \centering
    \includegraphics[width=\linewidth]{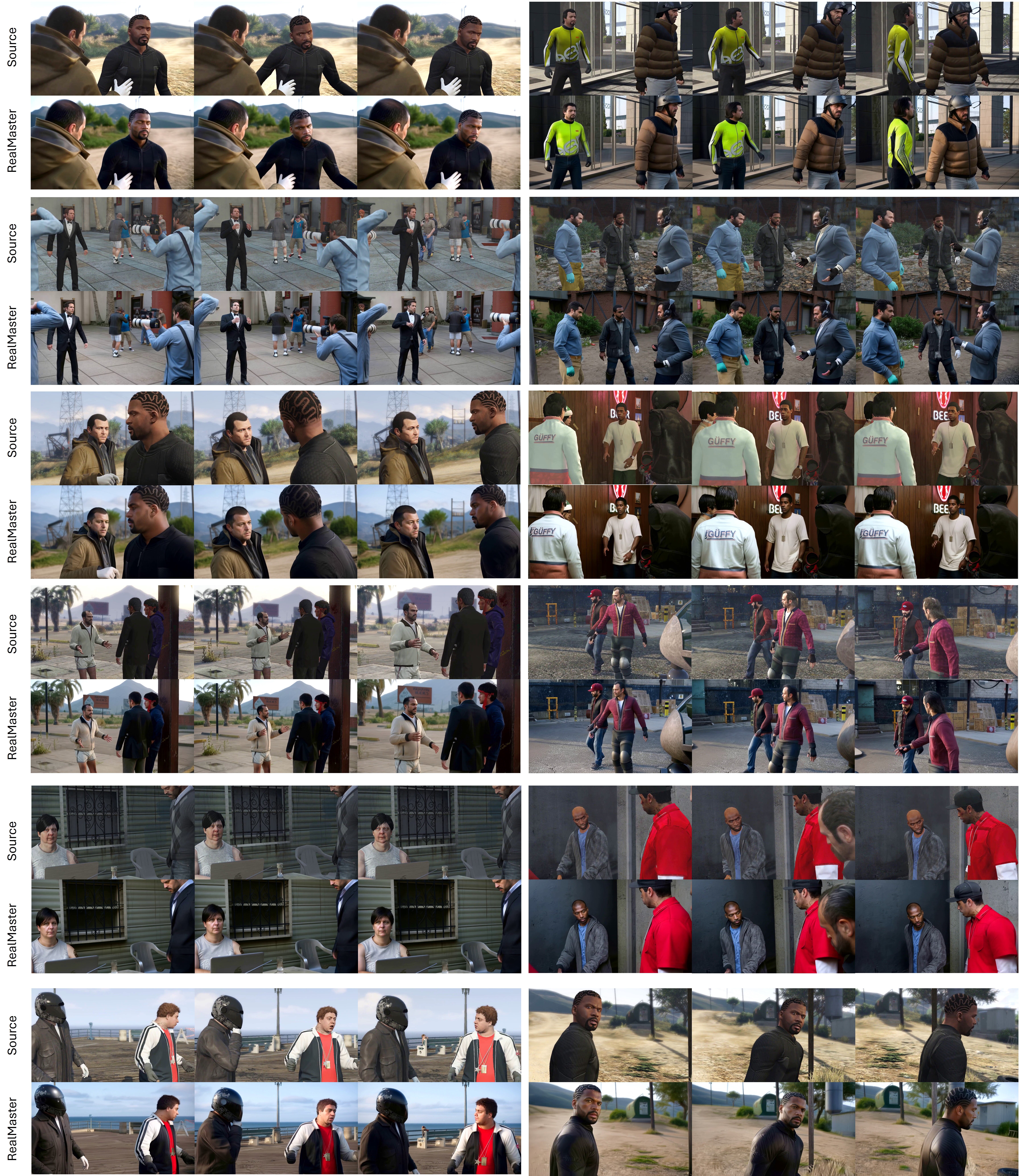}
    \caption{Additional Qualitative Results.}
    \label{fig:qualitative_results_figure_page_1}
\end{figure*}

%% file: figures/figure_page_2.tex
\begin{figure*}[t]
    \centering
    \includegraphics[width=\linewidth]{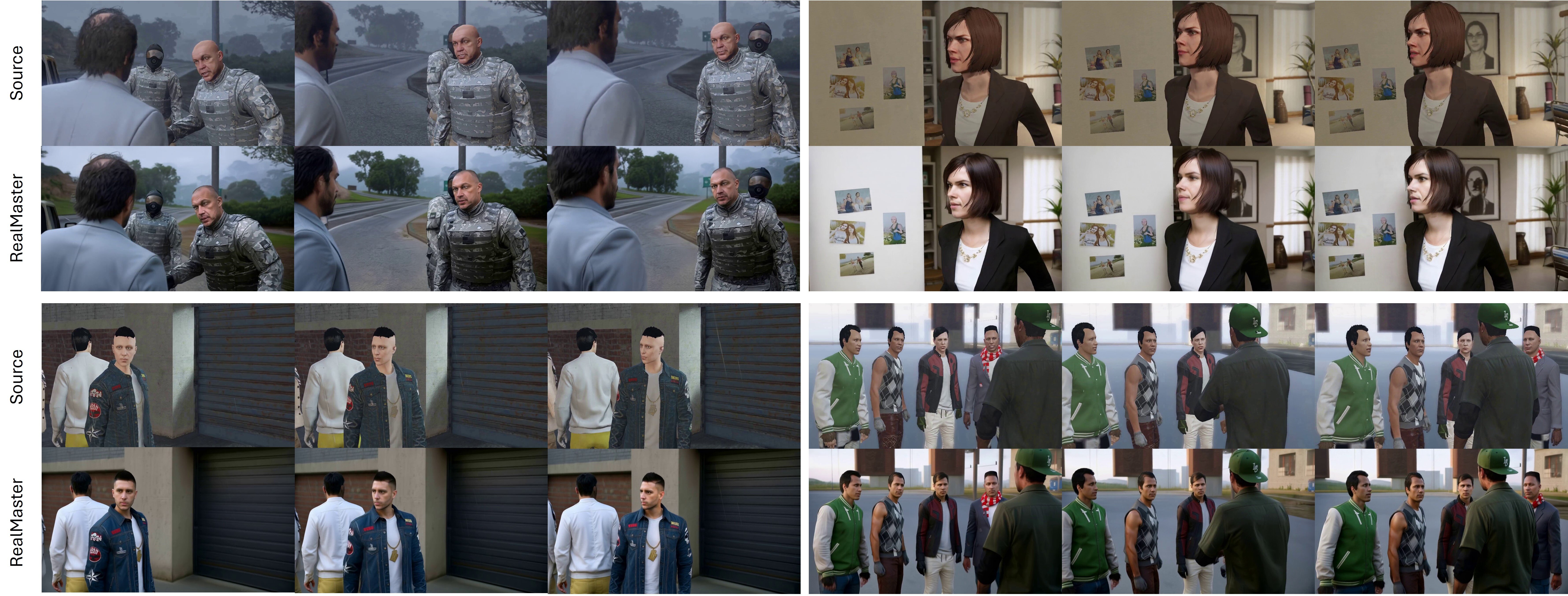}
    \caption{Additional Qualitative Results.}
    \label{fig:additional_qualitative_results_figure_page}
\end{figure*}

\begin{figure*}[!t]
    \centering
    \includegraphics[width=\textwidth]{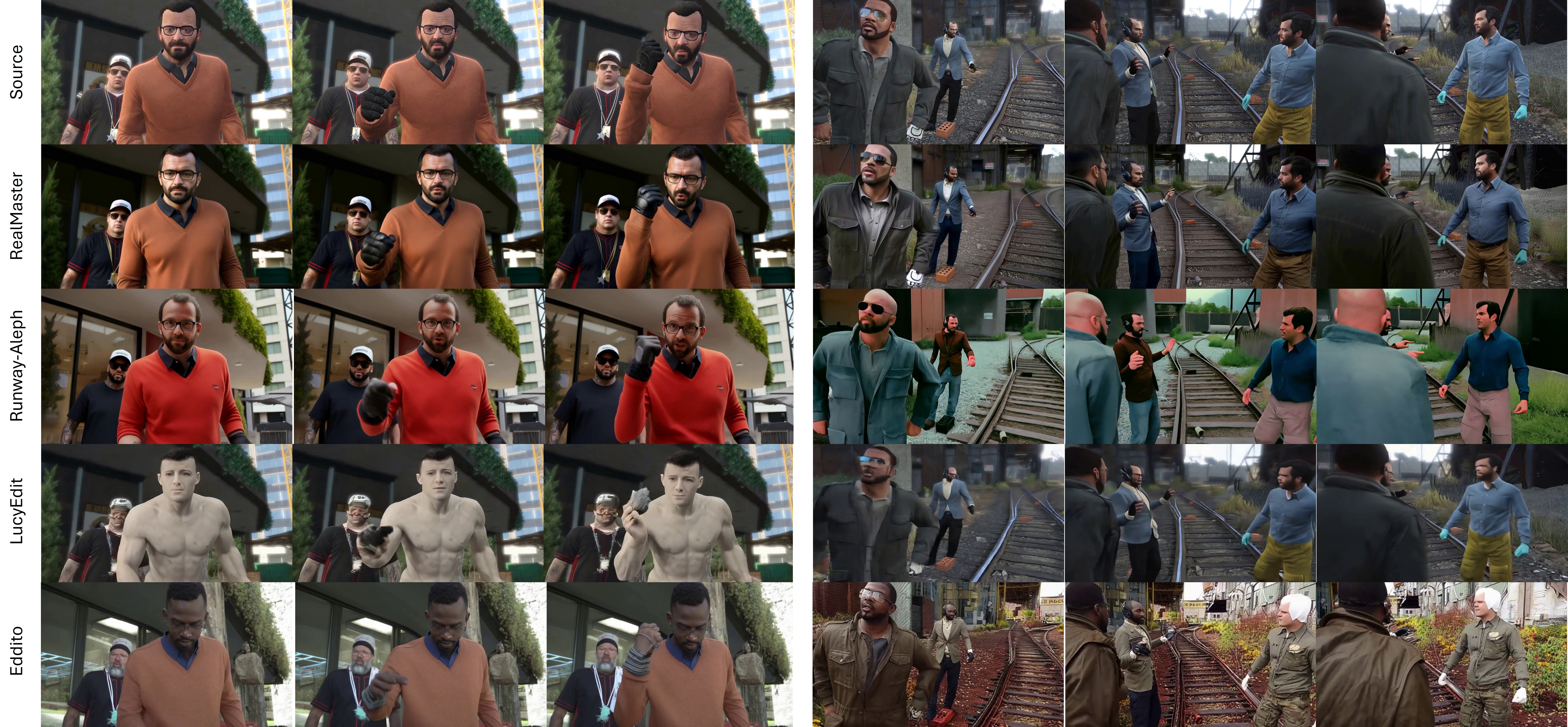}
    \caption{{Additional qualitative comparisons with baseline methods.}
    }
    \label{fig:comparison_figure_page}
\end{figure*}

\begin{figure*}[t]
  \centering
  \includegraphics[width=1\linewidth]{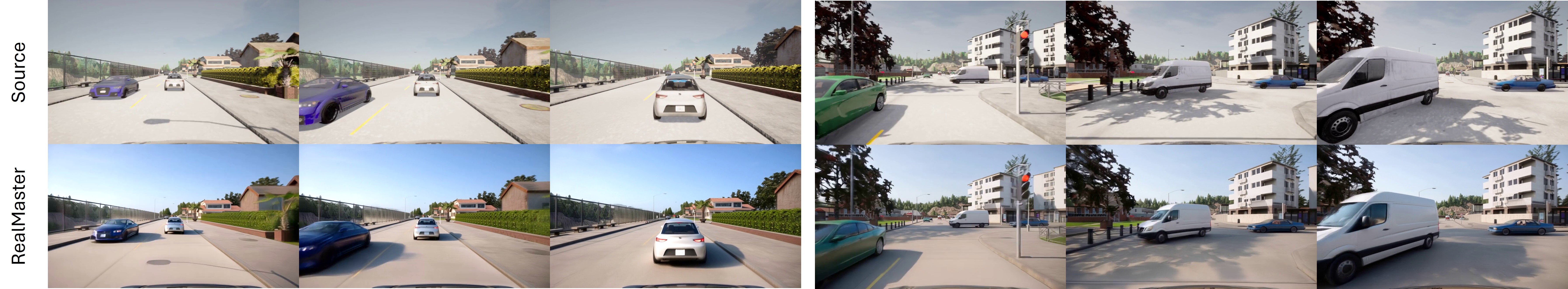}
    \caption{Additional generalization results on the CARLA-LOC dataset}
  \label{fig:additiona_carla}
\end{figure*}

%% file: supp.tex
\section{Failure Cases}
\label{sec:supp_failure_cases}

\begin{figure}[h]
  \centering
  \includegraphics[width=\linewidth]{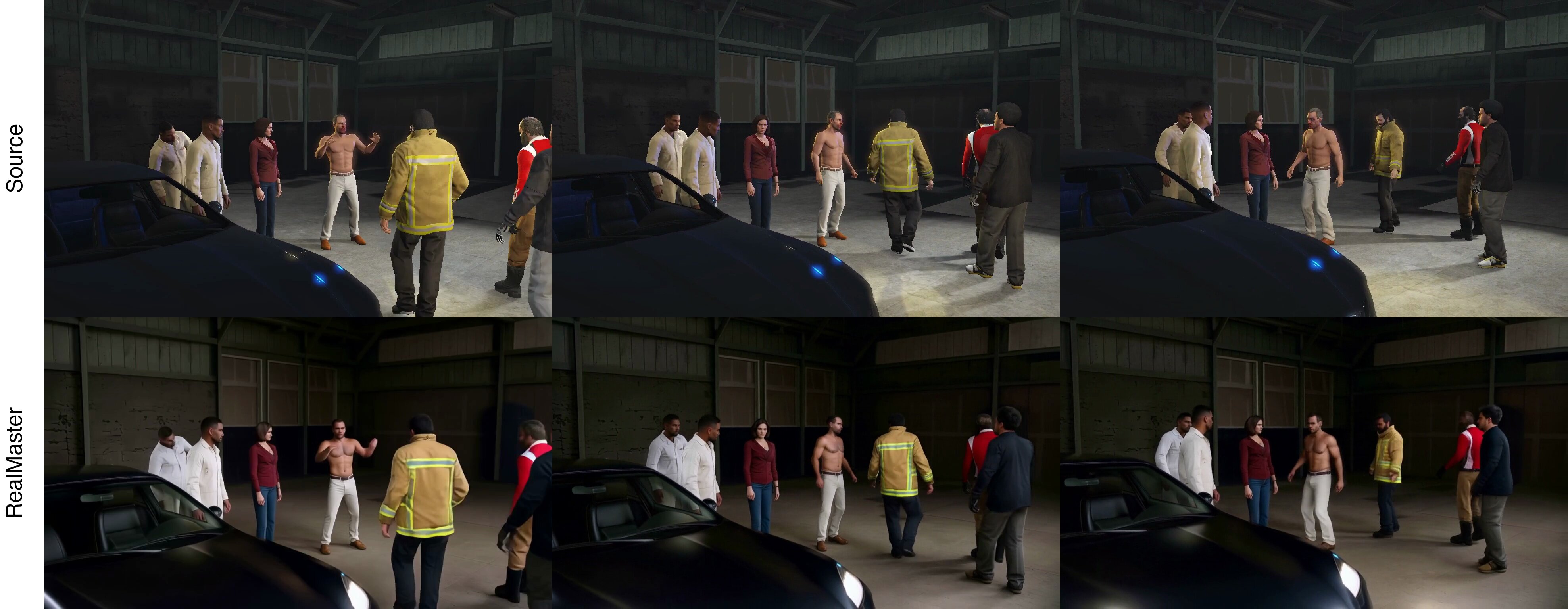}
  \vspace{0.5em}
  \includegraphics[width=\linewidth]{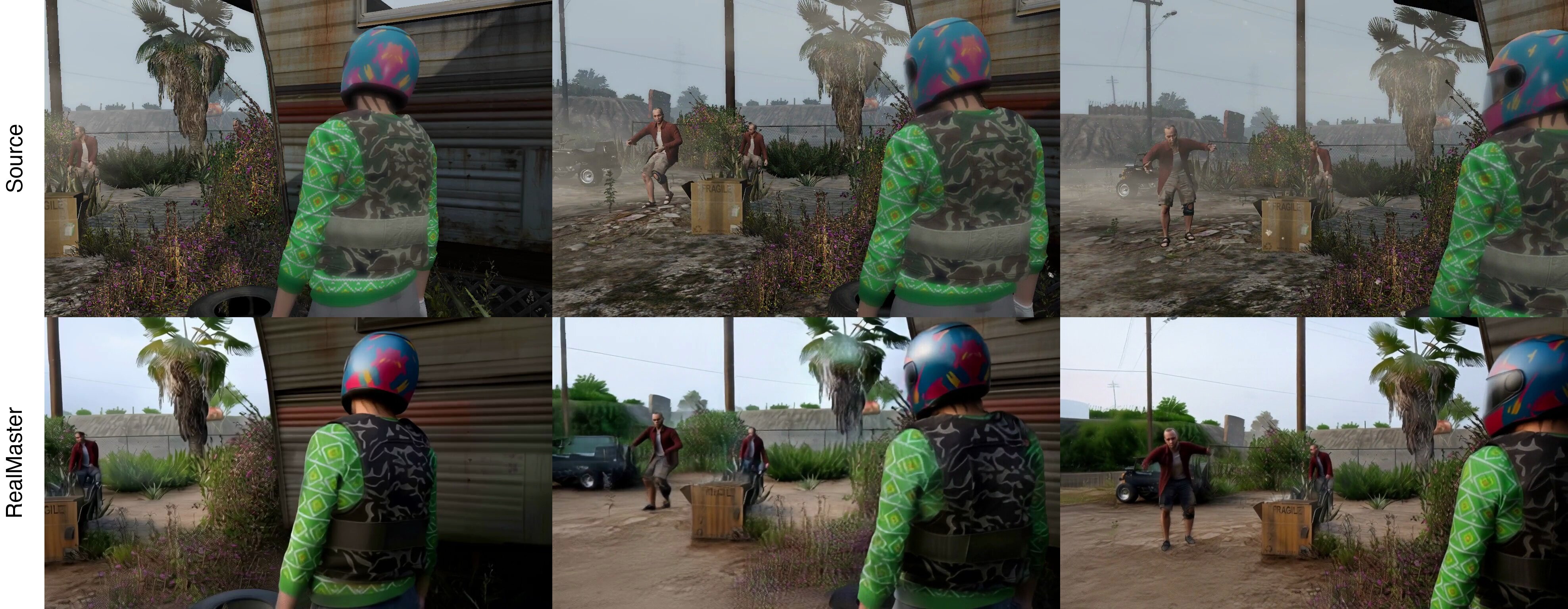}
  \vspace{-1em}
  \caption{\textbf{Failure cases.}
    Top: overly conservative output on a scene
    with small, distant objects.
    Bottom: temporal artifacts caused by fast
    camera and character motion.}
  \label{fig:failure_cases}
\end{figure}

We identify two main failure modes of RealMaster,
illustrated in \cref{fig:failure_cases}.
First, when the scene contains many small, distant
objects, the model tends to be overly conservative,
producing only subtle photorealistic changes that
are hard to notice at full-frame resolution.
This behavior is inherited from the image editing
model, Qwen-Image-Edit, used in the data generation
pipeline, which similarly struggles to enhance small
objects.
Second, scenes with fast camera or character motion
lead to temporal artifacts in the output.
This limitation is inherited from the base video
diffusion model, which was not designed to
handle large inter-frame displacements.

\section{Additional Implementation Details}
\label{sec:supp_impl_details}

This section provides additional implementation details for the data
generation pipeline and for LoRA training.

\subsection{Data generation pipeline}
\label{subsec:supp_datagen}

We sample 81-frame clips from the SAIL-VOS training set and upsample
videos from 8\,fps to 16\,fps by repeating each frame at
$800 \times 1200$ resolution.
For each clip, we edit the first and last frames with
Qwen-Image-Edit using the prompt
\texttt{"make it look photorealistic"}
and treat them as appearance anchors.
We then use VACE to propagate anchor appearance to the
intermediate frames, conditioned on an edge representation
of each input frame.

To improve identity consistency, we filter generated pairs
using ArcFace.
We retain a clip only if the mean ArcFace cosine similarity
between detected faces in the rendered input and in the
generated output exceeds 0.4.
Out of 3,050 initial clips, this filtering retains 1,216
training clips, removing approximately 60\% of the
generated data.

\subsection{Model training}
\label{subsec:supp_training}

We fine-tune Wan2.2 T2V-A14B using an IC-LoRA training setup.
We encode the rendered input clip as clean reference tokens with the timestep
fixed to $t{=}0$, and share positional encoding with the noisy target tokens
that are denoised toward the photorealistic target.

We provide a detailed summary of the hyperparameters used for training
RealMaster in \cref{tab:hyperparameters}.

\input{tables/hyperparams}

\subsection{Baseline configurations}
\label{subsec:supp_baselines}

For all three baselines, Runway-Aleph, LucyEdit, and Editto,
we use the prompt
\texttt{"make the video look photorealistic"}.
All other settings follow the default configurations provided
by the respective authors.

\section{Evaluation Metric Details}
\label{sec:supp_metrics}

\subsection{GPT-RS}
\label{subsec:supp_gptrs}

We use GPT-4o as a rubric-based judge to rate photorealism.
We report two variants.
GPT-RS$_{\text{with-ref}}$ provides the rendered input frame as a reference and
asks the judge to consider both photorealism and faithfulness.
GPT-RS$_{\text{no-ref}}$ provides only the edited frame and asks the judge to
score photorealism only.
In both cases, the model returns valid JSON with a single integer key
\texttt{rating} in the range 1 to 10.

\paragraph{\textbf{GPT-RS$_{\text{with-ref}}$ system prompt.}}
\begin{lstlisting}
You are an expert evaluator of GTA-to-photoreal image translation.
You will be shown TWO images:
1) The original GTA game frame
2) The edited image produced by a model attempting photorealism
Your task:
Evaluate how successful the edited image is as a faithful, photorealistic
transformation of the original GTA frame.

Faithfulness requirements:
- Same scene layout and camera viewpoint
- Same object positions, object colors and proportions
- No hallucinated, removed, or swapped objects
- No major geometric changes (bending, drifting, resizing)

Photorealism focus:
- Geometry stability (warping, melting, bending)
- Lighting and shadows (direction, contact, consistency)
- Materials and textures (plastic look, over-smoothing, repetition)
- Fine detail (grain, sharpness, depth of field)
- Text and signage (legible, stable, non-gibberish)
- Neural artifacts (halos, ghosting, ringing, checkerboard)

Score the QUALITY OF THE EDIT, considering BOTH:
1) Faithfulness to the original GTA image
2) Photorealism of the edited image
Scale (1-10):
10 = Faithful and indistinguishable from real camera footage
8-9 = Faithful with minor realism flaws visible on close inspection
6-7 = Mostly faithful; noticeable synthetic artifacts or small inconsistencies
4-5 = Partially faithful; clear mismatches or strong realism artifacts
1-3 = Unfaithful or failed transformation (hallucinations, scene changes, or
      severe artifacts)

Return valid JSON only with a single key: rating (integer 1-10).
\end{lstlisting}

\paragraph{\textbf{GPT-RS$_{\text{with-ref}}$ user prompt.}}
\begin{lstlisting}
Image 1 is the original GTA frame.
Image 2 is the edited output attempting photorealism.
Rate how successful the edit is.
\end{lstlisting}

\paragraph{\textbf{GPT-RS$_{\text{no-ref}}$ system prompt.}}
\begin{lstlisting}
You are an expert in video synthesis results.
Your task is to judge whether the provided image could plausibly be a real
camera frame.
Score photorealism only. Ignore aesthetics or artistic quality.
Scale (1-10, photorealism):
10 = Indistinguishable from real camera footage
8-9 = Looks real at a glance; minor flaws on close inspection
6-7 = Mixed realism; noticeable synthetic artifacts but partially plausible
4-5 = Clearly synthetic; CG or translation artifacts are obvious, but image is
      coherent
1-3 = Obviously synthetic or broken; severe artifacts make it unmistakably fake
Focus on:
- Geometry stability (warping, melting, bending)
- Lighting and shadows (direction, contact, consistency)
- Materials and textures (plastic look, over-smoothing, repetition)
- Fine detail (grain, sharpness, depth of field)
- Text and signage (legible, stable, non-gibberish)
- Neural artifacts (halos, ghosting, ringing, checkerboard)
Be conservative: if multiple strong artifacts exist, do not score above 7.
Return valid JSON only with a single key: rating (integer 1-10).
\end{lstlisting}

\paragraph{\textbf{GPT-RS$_{\text{no-ref}}$ user prompt.}}
\begin{lstlisting}
This image is a frame produced by a model that converts GTA gameplay into
realistic video. Rate how realistic it looks as real camera footage.
\end{lstlisting}

%% file: tables/hyperparams.tex
\begin{table}[t]
    \centering
    \caption{\textbf{Training Hyperparameters.} Summary of the configuration
    used for fine-tuning RealMaster.}
    \label{tab:hyperparameters}
    \small
    \begin{tabular}{lc}
        \toprule
        \textbf{Hyperparameter} & \textbf{Value} \\
        \midrule
        Base Model & Wan2.2 T2V-A14B \\
        LoRA Rank & 32 \\
        Optimizer & AdamW \\
        Learning Rate & $1 \times 10^{-4}$ \\
        Batch Size & 8 \\
        Total Training Steps & 1,200 \\
        Resolution & $800 \times 1200$ \\
        Frames per Clip & 81 \\
        Training Hardware & 1 $\times$ H200 \\
        \bottomrule
    \end{tabular}
\end{table}